\documentclass{article}

\PassOptionsToPackage{numbers, compress, sort}{natbib}

\usepackage[final]{neurips_2025}

\PassOptionsToPackage{table}{xcolor}
\usepackage{xcolor}
\usepackage[utf8]{inputenc} 
\usepackage[T1]{fontenc}    
\usepackage{hyperref}       
\usepackage{url}            
\usepackage{booktabs}       
\usepackage{amsfonts}       
\usepackage{nicefrac}       
\usepackage{microtype}      
\usepackage[pdftex]{graphicx}
\usepackage{pifont}
\usepackage{caption}
\usepackage{makecell}
\newcommand{\cmark}{\ding{51}}%
\newcommand{\xmark}{\ding{55}}%
\usepackage{svg}
\usepackage{ctable}
\usepackage[normalem]{ulem}
\usepackage{amsmath}
\usepackage[noabbrev,nameinlink]{cleveref} 
\usepackage{tabularx} 
\usepackage{array}    
\usepackage{subcaption}

\usepackage{arydshln}
\usepackage{enumitem}
\usepackage{multirow}
\usepackage{float}
\usepackage{wrapfig}
\usepackage[nolist,nohyperlinks]{acronym}
\usepackage{algorithm, algpseudocode}
\algrenewcommand\algorithmicrequire{\textbf{Input:}}
\algrenewcommand\algorithmicensure{\textbf{Output:}}
\usepackage{siunitx}
\usepackage{algorithm}
\usepackage{algpseudocode}
\usepackage{setspace} 

\crefname{figure}{fig.}{figs.}
\Crefname{figure}{Fig.}{Figs.}

\crefname{table}{tab.}{tabs.}
\Crefname{table}{Tab.}{Tabs.}

\crefname{section}{sec.}{secs.}
\Crefname{section}{Sec.}{Secs.}

\crefname{appendix}{app.}{apps.}
\Crefname{appendix}{App.}{Apps.}

\newcolumntype{L}[1]{>{\raggedright\arraybackslash}p{#1\linewidth}}
\newcolumntype{C}[1]{>{\centering\arraybackslash}p{#1\linewidth}}
\newcolumntype{R}[1]{>{\raggedleft\arraybackslash}p{#1\linewidth}}

\title{FOCUS: Internal MLLM Representations for\\ Efficient Fine-Grained Visual Question Answering}

\begin{acronym}
\acro{MLLM}[MLLM]{Multimodal Large Language Model}
\acro{LLM}[LLM]{Large Language Model}
\acro{VQA}[VQA]{Visual Question Answering}
\acro{FP}[FP]{Forward Pass}
\acro{GT}[GT]{Ground Truth}
\acro{ROI}[ROI]{region of interest}
\acro{NMS}[NMS]{Non-Maximum Suppression}
\acro{FOCUS}[FOCUS]{Fine-grained visual Object Cropping Using cached token Similarity}
\end{acronym}

\acrodefplural{ROI}[ROIs]{regions of interest}
\acrodefplural{FP}[FPs]{Forward Passes}

\newcommand{\rebuttal}[1]{\textcolor{black}{#1}}

\author{Liangyu Zhong$^{* 1,3}$, 
Fabio Rosenthal$^{* 2,4}$,
Joachim Sicking\hspace{1pt}$^{3}$,
Fabian Hüger\hspace{1pt}$^{3}$,\\
\textbf{
Thorsten Bagdonat\hspace{1pt}$^{4}$, 
Hanno Gottschalk\hspace{1pt}$^{1}$,
Leo Schwinn\hspace{1pt}$^{2}$
}\\[2mm]
$^1$\hspace{1pt}Technical University of Berlin, $^2$\hspace{1pt}Technical University of Munich,\\
$^3$\hspace{1pt}CARIAD SE, $^4$\hspace{1pt}Volkswagen AG \\[3mm]
\textbf{Project page:} {\color{magenta}\href{https://focus-mllm-vqa.github.io/}{https://focus-mllm-vqa.github.io}}
}

\begin{document}

\maketitle

\begin{abstract}
  While \acp{MLLM} offer strong perception and reasoning capabilities for image-text input, Visual Question \mbox{Answering (VQA)} focusing on small image details still remains a challenge. 
  Although visual cropping techniques seem promising, recent approaches have several \mbox{limitations: the need} for task-specific fine-tuning, low efficiency due to uninformed exhaustive search, or incompatibility with efficient attention implementations.
  We address these shortcomings by proposing a training-free visual cropping method, dubbed \texttt{FOCUS}, that leverages \ac{MLLM}-internal representations to guide the search for the most relevant image region.
  This is accomplished in four steps: first, we identify the target object(s) in the VQA prompt; second, we compute an object \mbox{relevance} map using the key-value (KV) cache; third, we propose and rank \mbox{relevant} \mbox{image} regions based on the map; and finally, we perform the fine-grained VQA task using the top-ranked region.
  As a result of this informed search strategy, \texttt{FOCUS} achieves strong performance across four fine-grained VQA datasets and three types of \acp{MLLM}.
  It outperforms three popular visual cropping methods in both accuracy and efficiency, and matches the best-performing baseline, ZoomEye, while requiring $3\,\text{--}\,6.5\times$ \mbox{less compute}.
\end{abstract}

{\renewcommand{\thefootnote}{}\footnotetext{*\,Equal contribution}}%

\section{Introduction}
\vspace{-0.5\baselineskip}
\label{sec:introduction}

\acfp{MLLM} exhibit compelling cross-modal perception and reasoning capabilities, particularly on image-text data \cite{instruct_blip_dai, llava_ov_li, intern_vl_chen}.
However, standard \ac{MLLM} architectures are not well suited to perceive and reason about small visual details in high-resolution images \cite{dc2_wang2024, vicrop_zhang2025} as they typically downscale their inputs, leading to a loss of information.
Examples of these so-called \textit{global-view} \acp{MLLM} include Instruct-BLIP \cite{instruct_blip_dai} and LLaVA-1.5 \cite{llava_15_liu}, which only support low-resolution inputs of $224 \times 224$~px and $336 \times 336$~px, respectively.
As a consequence, global-view \acp{MLLM} perform poorly on \acf{VQA} tasks involving small-scale visual~details \cite{vstar_wu24}.

Recent \ac{MLLM} architectures such as LLaVA-OneVision \cite{llava_ov_li} and Gemma-3 \cite{gemma_3_kamath} address this limitation by processing both a downsampled global view and local crops extracted from the original image.
This dual-view strategy enables them to handle high-resolution inputs with reduced information loss compared to global-view \acp{MLLM}.
However, despite having access to fine-grained visual details from all local crops, these so-called \textit{global-local-view} \acp{MLLM} struggle to identify the few visual tokens that are relevant for fine-grained \ac{VQA} amid the large number of local crop tokens.
While these global-local-view architectures outperform global-view \acp{MLLM}, their effectiveness on fine-grained \ac{VQA} tasks still remains limited \cite{dc2_wang2024}.

An orthogonal research direction to address the limitations of \acp{MLLM} in capturing fine details in high-resolution images are visual cropping approaches \cite{dc2_wang2024, vicrop_zhang2025, vstar_wu24, zoomeye_shen2024}, which seek to pass only relevant image regions to the \ac{MLLM}.
However, popular visual cropping techniques like SEAL \cite{vstar_wu24}, DC\textsuperscript{2} \cite{dc2_wang2024}, ZoomEye \cite{zoomeye_shen2024}, and ViCrop \cite{vicrop_zhang2025} suffer from one or more of the following limitations: (1) reliance on task-specific fine-tuning of \acp{MLLM} for fine-grained \ac{VQA}, (2) use of inefficient, exhaustive search algorithms, and (3) incompatibility with efficient attention implementations such as FlashAttention \cite{fa_dao} (see \Cref{tab:comparison} and \Cref{fig:comparison_baseline_methods} for a visual comparison of the methods). 
We propose a visual cropping method, termed \textbf{F}ine-grained visual \textbf{O}bject \textbf{C}ropping \textbf{U}sing cached token \textbf{S}imilarity \texttt{(FOCUS)}, that addresses these issues as is outlined in the following.

\begin{wraptable}{r}{0.5\textwidth}
    \vspace{-13pt}
    \caption{\textbf{Comparison of visual cropping \mbox{methods} w.r.t.~desirable properties.} Unlike previously suggested methods, \texttt{FOCUS} is training-free, algorithmically efficient in search, and compatible with efficient attention implementations.}
\label{tab:comparison}
\renewcommand{\arraystretch}{1.2}
\resizebox{\linewidth}{!}{
    \begin{tabular}{ l c c c}
      \toprule
      \textbf{Method} & \textbf{\makecell{Training-\\free}} & \textbf{\makecell{Efficient \\ search algo.}} & \textbf{\makecell{Compatible w/ \\ efficient attention}}\\
      \midrule
      SEAL \cite{vstar_wu24} & \xmark & \xmark & \cmark \\
      DC\textsuperscript{2} \cite{dc2_wang2024} & \cmark & \xmark & \cmark \\
      ZoomEye \cite{zoomeye_shen2024} & \cmark & \xmark & \cmark \\
      ViCrop \cite{vicrop_zhang2025} & \cmark & \cmark & \xmark \\
      \midrule
      \texttt{FOCUS} (Ours) & \cmark & \cmark & \cmark \\
      \bottomrule
    \end{tabular}
}

\end{wraptable}
To tackle limitation (1), \texttt{FOCUS} leverages the internal representations of \acp{MLLM}, specifically their key-value (KV) caches \cite{kvcache_pope}, to localize question-relevant image regions in a training-free manner—unlike the SEAL \cite{vstar_wu24} technique.
Moreover, to mitigate limitation (2), our method includes textual clues to enable object-aware \mbox{localization} without exhaustive cropping of the image, thereby improving the algorithmic efficiency—different from DC\textsuperscript{2} \cite{dc2_wang2024} and ZoomEye \cite{zoomeye_shen2024}.
To overcome limitation (3), \texttt{FOCUS} utilizes the cached value features readily available during inference, making it natively compatible with efficient attention implementations \cite{fa_dao}—unlike ViCrop \cite{vicrop_zhang2025} that depends on full attention weights.
Specifically, \texttt{FOCUS} combines these components as follows: for each VQA question, we first identify the target object(s) in the question prompt.
Second, we construct an object relevance map using cosine similarity between the cached text tokens of the target object(s) and the cached image tokens, and then propose relevant regions based on this map.
Third, we rank the proposed image regions based on the existence confidence of the target object in each region.
Finally, we perform \ac{VQA} solely based on the image region with the highest confidence. Note that \texttt{FOCUS} is compatible with both global- and global-local-view \acp{MLLM}. 

We evaluate \texttt{FOCUS} on the fine-grained VQA datasets V*Bench \cite{vstar_wu24}, HRBench-4K \cite{dc2_wang2024}, HRBench-8K \cite{dc2_wang2024} and MME-RealWorld-Lite \cite{mmerealworld_zhang2025}.
Across the first three datasets, our method achieves on average $42\%$ higher accuracy over the vanilla \acp{MLLM} when using LLaVA-1.5 and $17\%$ when using LLaVA-OneVision, while improving LLaVA-OneVision by $6\%$ on the multi-domain MME-RealWorld-Lite dataset.
Moreover, \texttt{FOCUS} achieves comparable or superior performance w.r.t.~the state-of-the-art baseline ZoomEye \cite{zoomeye_shen2024} while being $3.5\,\text{--}\,4.5\times$ more efficient with LLaVA-1.5 and $3\,\text{--}\,6.5\times$ more efficient with LLaVA-OneVision.

Our key contributions are as follows: 
First, we propose \texttt{FOCUS}, a training-free visual cropping method for \ac{MLLM}-based fine-grained \ac{VQA} that identifies relevant image regions using internal representations of the \ac{MLLM}.
Second, we provide extensive empirical evidence for \texttt{FOCUS}'s favorable accuracy-efficiency trade-offs compared to previous visual cropping methods for fine-grained \ac{VQA}.
Third, we conduct an ablation study that provides insights on how \texttt{FOCUS} leverages \ac{MLLM}-internal knowledge for efficient visual cropping.

\section{Related work}
\label{sect:related_work}
\vspace{-0.5\baselineskip}

\ac{VQA} involves answering a question based on the visual content of an image.
While various types of machine learning models can be applied to this task, \acp{MLLM} have become the de~facto standard due to their strong cross-modal reasoning capabilities \cite{llava_15_liu, vqa_v2_goyal}.
Here, we focus on the multiple-choice variant of \ac{VQA}, where the \ac{MLLM} is expected to select the correct answer from a fixed set of options.
In the following paragraphs, we describe the datasets commonly used to evaluate \ac{MLLM} performance on \textit{fine-grained \ac{VQA}} tasks, i.e., tasks that require focus on visual details.
Further, we provide a technical overview of recent visual cropping methods.

\begin{wrapfigure}{l}{0.5\textwidth}
  \vspace{13.5pt}
  \centering
  \includegraphics[width=0.45\textwidth]{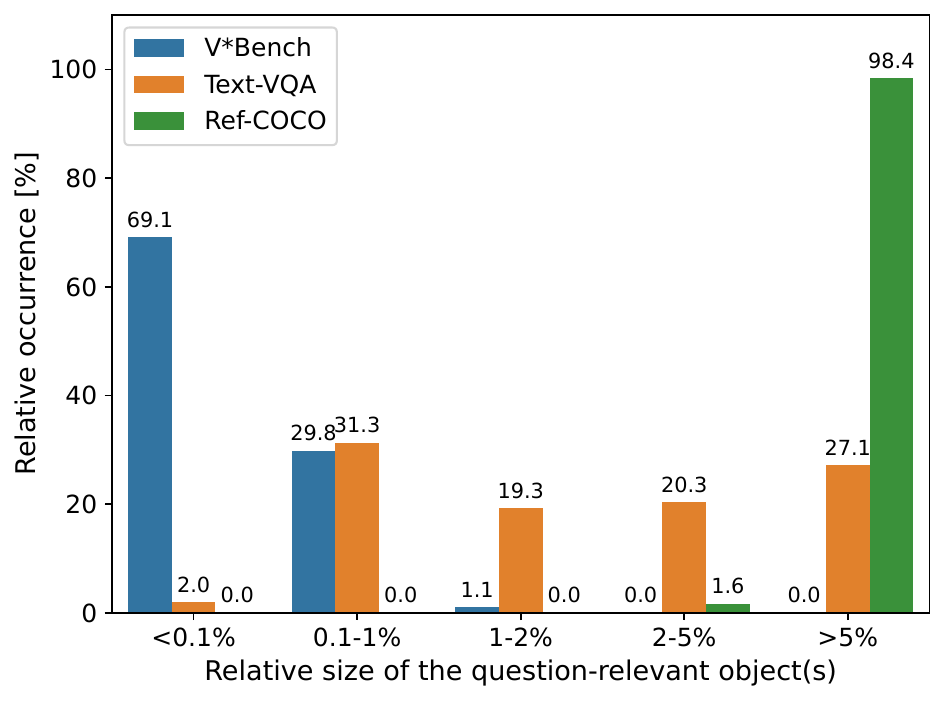}
  \caption{\textbf{Many VQA datasets focus on large instead of tiny objects.} This figure shows the relative area of question-relevant objects w.r.t.\;the image. V*Bench contains various tiny \ac{VQA}-relevant objects.}
  \label{fig:gt_area_datasets}
  \vspace{-6.5pt}
\end{wrapfigure}

\textbf{Fine-grained \ac{VQA} datasets}\hspace{1.8mm}
In common \ac{VQA} datasets such as Text-VQA \cite{textvqa_singh} and \mbox{Ref-COCO}~\cite{ref_coco_kazemzadeh}, the relevant objects are typically prominent within the image.
In contrast, fine-grained \ac{VQA} tasks focus on much smaller visual targets.
The V*Bench dataset exemplifies this, containing significantly smaller question-relevant objects compared to the aforementioned datasets (see \Cref{fig:gt_area_datasets}).
Additional fine-grained \ac{VQA} datasets include \mbox{HRBench-4K}~\cite{dc2_wang2024}, HRBench-8K \cite{dc2_wang2024}, and MME-RealWorld-Lite \cite{mmerealworld_zhang2025}.
Among these, V*Bench is the only dataset that provides \ac{GT} annotations for question-relevant objects.

\textbf{Visual cropping for fine-grained \ac{VQA}}\hspace{1.8mm}
In this emerging area of research, different visual cropping methods \cite{zoomeye_shen2024, dc2_wang2024, vstar_wu24, vicrop_zhang2025} have been proposed to improve \ac{MLLM} performance on fine-grained \ac{VQA} tasks (see \Cref{tab:comparison}).
SEAL \cite{vstar_wu24} employs a dual-\ac{MLLM} setup: one MLLM for visual search and another one for the actual \ac{VQA} task.
The visual search model with additional decoders is fine-tuned to predict object heatmaps and coordinates.
It performs a top-down hierarchical search, generating contextual cues to iteratively locate the target object based on confidence scores.
This approach requires task-specific fine-tuning and is rather inefficient due to its complex, multi-module design and recursive search process.
In contrast, DC\textsuperscript{2} \cite{dc2_wang2024} constructs a hierarchical image region tree from the global view down to regions, matching the base resolution of the vision encoder.
Each region is captioned using the \ac{MLLM} and those containing the target object are merged for actual \ac{VQA}. 
While training-free, DC\textsuperscript{2} is inefficient due to the extensive tree traversal and costly captioning process.
Similar to DC\textsuperscript{2}, ZoomEye \cite{zoomeye_shen2024} employs a hierarchical tree search, but instead of captioning, it predicts a confidence score for each image region.
This score is computed through a complex mechanism involving three sequential \ac{MLLM} forward passes, each using a different question prompt.
As a result, the process cannot be simplified or shared across regions, making the method inefficient due to both the hierarchical search and the high cost of confidence estimation.
ViCrop \cite{vicrop_zhang2025} is an efficient, training-free visual cropping method that avoids hierarchical search by directly computing a question-guided heatmap to localize the target object. 
However, its best-performing variants depend on Q-K attention weights and answer gradients, making the method incompatible with efficient attention implementations such as FlashAttention \cite{fa_dao}.
Unlike these methods, our method is a training-free visual cropping approach without additional modules besides the \ac{MLLM}.
Rather than relying on recursive search, captioning, or Q-K attention weights, we employ an informed search guided by internal representations of the \ac{MLLM} that is compatible with efficient attention implementations.
This design enables our method to achieve significantly higher efficiency without sacrificing accuracy.

\section{FOCUS: Fine-grained visual object cropping using cached token similarity}\label{sect:method}
\vspace{-0.5\baselineskip}

\begin{figure}[h]
    \centering
    \includegraphics[width=1\textwidth]{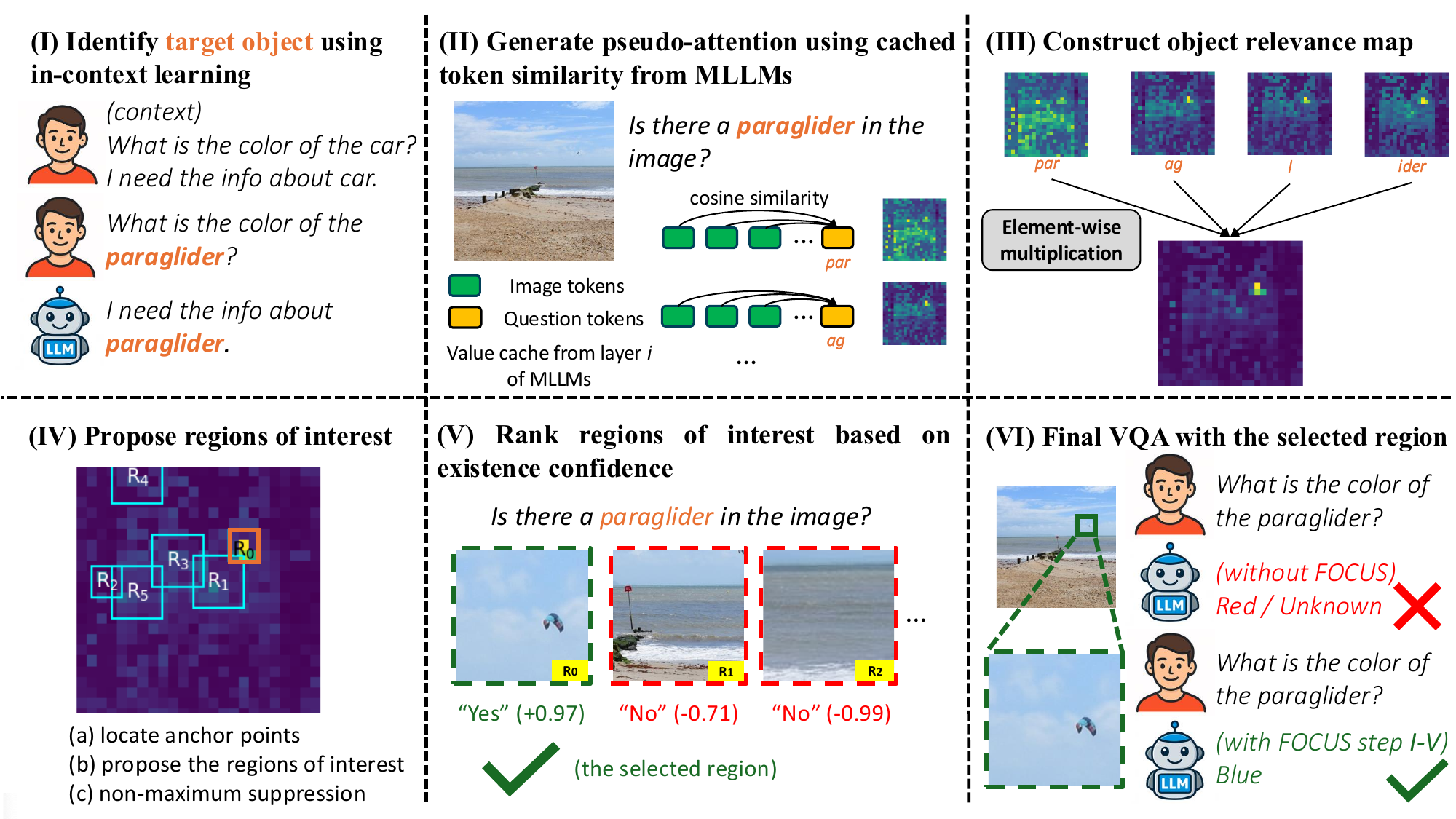}
    \caption{\textbf{Overview of \texttt{FOCUS}}. The method identifies the target objects mentioned in the question \textbf{(I)} and constructs their object relevance maps using cosine similarity between cached tokens \textbf{(II + III)}. Then, it proposes regions of interest and ranks those by the existence confidence of the target object in each region \textbf{(IV + V)}. Finally, the selected region is used to perform \ac{VQA} \textbf{(VI)}.}
    \label{fig:flow_chart}
\end{figure}

We first provide relevant background information for our method in \Cref{subsect:background}.
Next, we describe in detail how our method proposes relevant image regions based on the KV-cache in \Cref{subsect:rel_map}.
Finally, we explain how these image regions are used for fine-grained \ac{VQA} tasks in \Cref{subsect:proposal_ranking_roi}.
We provide a visualization of \texttt{FOCUS} in \Cref{fig:flow_chart}.

\subsection{Background}
\label{subsect:background}
\vspace{-0.3\baselineskip}

\acp{MLLM} typically comprise three core components: a vision encoder, a modality projector, and a \ac{LLM}. The \ac{LLM} receives an input sequence of tokens $\mathrm{X} = (x_1, \ldots, x_n)$ and predicts the next token $x_{n+1}$ auto-regressively \cite{gpt1_radford}.
This sequence can be viewed as a concatenation of visual tokens $\mathrm{X}_{\text{vis}} = (x_1, \ldots, x_{a^2})$ and instruction textual tokens $\mathrm{X}_{\text{text}} = (x_{a^2+1}, \ldots, x_n)$, where $a$ is the grid size of the vision encoder (for simplicity, we ignore system prompt tokens here). 
In many models, the visual tokens from the vision encoder are extracted from a downscaled global view of the image and projected into the \ac{LLM}’s feature space via the modality projector.
The resizing aligns the input image with the base resolution expected by the vision encoder, e.g., $336 \times 336$ for CLIP \cite{clip_radford} and $384 \times 384$ for SigLIP \cite{siglip_zhai23}.
While this regime is effective in many scenarios, it tends to fail on fine-grained \ac{VQA} tasks \cite{dc2_wang2024, vicrop_zhang2025}.
To address this limitation, many global-local-view \acp{MLLM} \cite{llava_uhd_guo, llava_ov_li, llava_next_li, minicpm_yao} partition the original (unresized) image into $b$ local crops in addition to the resized global view.
These crops are encoded into additional visual tokens, extending the visual input to $\mathrm{X}_{\text{vis}} = (x_1, \ldots, x_{a^2 \cdot (b+1)})$.
This results in a rapid growth of computational cost due to the quadratic complexity of self-attention in transformer layers \cite{transformer_vaswani}.
We aim to improve the performance of both global-view and global-local-view \acp{MLLM} on fine-grained \ac{VQA} tasks by constructing a map that efficiently localizes the image regions most relevant to a target object mentioned in a question. This map is referred to as the \texttt{object relevance map}.
Once the relevant regions are identified, we restrict processing to these areas for the final \ac{VQA} prediction, thereby improving accuracy and computational efficiency.

\subsection{Constructing object relevance maps from cached token similarity} 
\vspace{-0.3\baselineskip}
\label{subsect:rel_map}

Localizing objects in images remains a challenge for \acp{MLLM} \cite{refl4_chen}.
While many models are fine-tuned on predicting bounding boxes \cite{ref_coco_kazemzadeh}, they often fail when prompted directly for the location of small objects, frequently producing hallucinated or imprecise coordinates \cite{refl4_chen}.
Instead of relying on explicit prompting, we propose to localize target objects by leveraging value features of cached tokens of \acp{MLLM}.
Recent work \cite{interpret_neo} shows that visual tokens in the sequence largely preserve spatial correspondence to their originating image regions across transformer layers \cite{transformer_vaswani}.
In fine-grained \ac{VQA} datasets \cite{vstar_wu24, dc2_wang2024, mmerealworld_zhang2025}, questions typically involve one or more specific objects in the image, which we refer to as the target object(s).
Since the text tokens corresponding to these targets co-exist alongside visual tokens in the token sequence, we estimate an object relevance map by computing the cosine similarity between the text and visual tokens.

To construct this object relevance map, we first identify the text tokens corresponding to the target object(s) as shown in \Cref{fig:flow_chart} (\textbf{I}).
Inspired by ZoomEye \cite{zoomeye_shen2024}, we use the in-context learning (ICL) capability of \acp{MLLM} \cite{incontext_dong} to extract the target object(s) by providing a few examples in the prompt.
This might extract one or multiple target object(s) from the question. 
For each target object, we apply a generic prompt template "\texttt{Is there a \{target object\} in the image?}" to query the \ac{MLLM} alongside the image.
Due to the tokenization \cite{transformer_vaswani} of the \ac{MLLM}, the target object can result in a sequence of target text tokens \smash{$\hat{X}_\text{{tgt}} = (\hat{x}_\text{0}, \ldots, \hat{x}_\text{i}, \ldots, \hat{x}_\text{s})$}.
We calculate the cosine similarity between the target tokens and the visual tokens in the sequence to construct the object relevance map.
While one might consider using standard query-key (Q-K) attention weights \cite{transformer_vaswani} for this purpose, many recent \acp{MLLM} employ efficient attention implementations such as FlashAttention \cite{fa_dao}, which do not generate Q-K attention weights explicitly. 
We address this issue by using value features preserved in the KV-cache mechanism \cite{kvcache_pope}, which is commonly used to accelerate autoregressive inference by storing intermediate representations.
Leveraging this, we propose an alternative value-value (V-V) pseudo-attention approach as shown in \Cref{fig:flow_chart} (\textbf{II}).
For each target token $\hat{x}_i^l, i \in \{0, \ldots, s\}$ in the $l$-th layer, we compute a pseudo-attention map \smash{$\mathbf{A}_{i}^l \in \mathbb{R}^{a\times a}$} by measuring its cosine similarity ($\cos$) with the visual tokens $(x_{1}^l, \ldots, x_{a}^l)$, where value features are available via the KV-cache:
\begin{equation}
    \mathbf{A}_i^l = \cos(\hat{x}_i^l, x_{j}^l),\;\;\; \text{for}\ j = 1, \ldots, a^2
\end{equation}

\rebuttal{and reshape $\mathbf{A}_i^l$ into an $a \times a$ matrix.}
Alternatively, one can also compute $\mathbf{A}_i^l$ using key features, see \Cref{subsect: ablation} for details.
In preliminary experiments, we empirically find that the pseudo-attention map $\mathbf{A}_{i}^l$ from a single layer might be noisy.
Therefore, we aggregate maps from $l$-th layer to $L$-th layer using attention rollout \cite{attnrollout_abnar}, incorporating residual connections to better preserve information flow as~follows:
\begin{align}
    \mathbf{A}_i &= \sum_{k=l}^{L} (\mathbf{A}_i^l + \mathbf{I}) / 2\ ,
\end{align}
where $\mathbf{I}$ denotes the identity matrix.
We aggregate the pseudo-attention maps $\mathbf{A}_i$ for each individual target token $\hat{x}_i$ by element-wise multiplication to capture consensus as shown in \Cref{fig:flow_chart} \textbf{(III)}: 
\begin{equation}
    \mathbf{A} = \mathbf{A}_0 \odot \mathbf{A}_1 \odot \cdots \odot \mathbf{A}_s\ .
\end{equation}
This operation allows only regions that are consistently highlighted across all target tokens to remain prominent.
For example, the token \texttt{red} may highlight many red objects, but when combined with \texttt{car}, only regions corresponding to red cars will be emphasized.
We refer to $\mathbf{A}$ as the object relevance map corresponding to the target object.
A normalization of $\mathbf{A}$ is performed after every matrix addition and multiplication to ensure numerical stability.

For global-local-view \acp{MLLM}, instead of calculating cosine similarity between the target tokens and the visual tokens from the global view, we use visual tokens extracted from the local crops.
We empirically find that these local tokens can better capture fine-grained details.
We compute the pseudo-attention map \smash{$\mathbf{A}_i^l$} using local visual tokens:
\begin{equation}
    \mathbf{A}_i^l = \cos(\hat{x}_i^l, x_{j}^l),\;\;\; \text{for}\ j = a^2+1, \ldots, a^2 \cdot (b+1)
\end{equation}

\rebuttal{and reshape $\mathbf{A}_i^l$ into a $h \times w$ matrix, where $h$ and $w$ denote the spatial dimensions of the local visual tokens arranged to closely preserve the original image's aspect ratio, so that $h\cdot w=a^2 \cdot b$.}
To reduce noise and enhance spatial coherence, we empirically apply a Gaussian filter to $\mathbf{A}$, followed by downsampling, yielding a cleaner object relevance map.

\subsection{Ranking of proposed \acp{ROI} for fine-grained \ac{VQA}}
\vspace{-0.3\baselineskip}
\label{subsect:proposal_ranking_roi}
Given an object relevance map, we define in the following a relevance score for each of its elements.
As shown in \Cref{fig:flow_chart} (\textbf{IV}), once the object relevance map is obtained, we extract the locations of the top-$k$ highest scores as anchor points on $\mathbf{A}$, which represent regions likely containing the target object.
To ensure sufficient spatial coverage, we select a relatively large 
$k$ and enforce a minimum distance $s_\text{dist}$ between anchor points. 
Anchor points that are too close are discarded.
Then, we generate an initial symmetric \ac{ROI} of minimal size $s_{\text{min}}$ per anchor point.

Each initial \ac{ROI} is then expanded up to the maximal size $s_{\text{max}}$, stopping when the average relevance score within the \ac{ROI} falls below a predefined threshold.
We rank the resulting \acp{ROI} based on the relevance score at their respective anchor points.
To eliminate redundancy, we apply non-maximum-suppression \cite{nms_hosang, r_cnn_girshick}, using a low threshold to promote diversity among selected regions.
This encourages broader spatial coverage.

The resulting object relevance map can be noisy due to spurious high-activation tokens \cite{register_darcet} that do not correspond to the target object.
As a result, a \ac{ROI} with a high relevance score may not actually contain the target object.
\rebuttal{Therefore, we verify whether the \ac{ROI} contains the target (see \Cref{subsct:final_vqa_infer} for details). Similar to ZoomEye \cite{zoomeye_shen2024}, the \ac{ROI} is provided to the \ac{MLLM} together with an existence prompt, and an existence confidence is computed from the model's response, as shown in \Cref{fig:flow_chart} (\textbf{V}).}
Then, we rerank the top $n_{\text{steps}}$ \acp{ROI} according to their existence confidence, with the number of steps $n_{\text{steps}}$ controlling the \ac{FP} budget.

\textbf{Final \ac{VQA} inference}\hspace{1.8mm}
In the previous paragraphs, we demonstrated how to obtain the most relevant \acp{ROI} for each target object.
These \acp{ROI} are now passed to the \ac{MLLM} for the final VQA prediction.
We follow the inference strategy from ZoomEye \cite{zoomeye_shen2024}, which categorizes questions in fine-grained \ac{VQA} datasets into two types.
\rebuttal{
Type-1 questions involve single-object instances and type-2 questions concern multiple instances of an object type.}
These categories can be automatically inferred using ICL or keyword-based heuristics, without requiring prior knowledge of the question.
For type-1 questions, we select the \ac{ROI} with the highest confidence score for each target object.
If the question involves multiple target objects, we combine the regions covering all relevant objects for the final \ac{VQA}.
For type-2 questions, we iterate over all proposed \acp{ROI} and select those with confidence scores higher than a threshold.
This step is not constrained by the $n_\text{steps}$.
In the case of global-local-view \acp{MLLM}, this process slightly differs.
Instead of combining the visual crops into a single area, we utilize the model's text-image-interleaved capabilities \cite{flamingo_alayrac, llava_next_li, llava_ov_li}.
Specifically, we provide one global image view—where all target objects are visually highlighted—alongside the top-confidence \acp{ROI} for each target object.
Please check \Cref{subsct:final_vqa_infer} for details.

\section{Experiments}\label{sect:experiments}
\vspace{-0.5\baselineskip}

We first describe the implementation details of our experiments in \Cref{subsect: impl_details} and provide the results on fine-grained \ac{VQA} datasets in \Cref{subsect:results}.
\rebuttal{Further, we conduct an ablation study and a hyperparameter sensitivity analysis in \Cref{subsect: ablation} and provide additional results (i) on \ac{VQA} datasets with larger objects and (ii) with Qwen-2.5-VL \cite{qwen_2_5_vl_bai} in \Cref{subsect: robustness_analysis}.}
Finally, we show some qualitative examples and discuss the limitations of \texttt{FOCUS} in \Cref{subsect:qualitative_examples} and \Cref{subsect:limitations}, respectively.

\begin{figure}[h]
    \centering
    \includegraphics[width=1\textwidth]{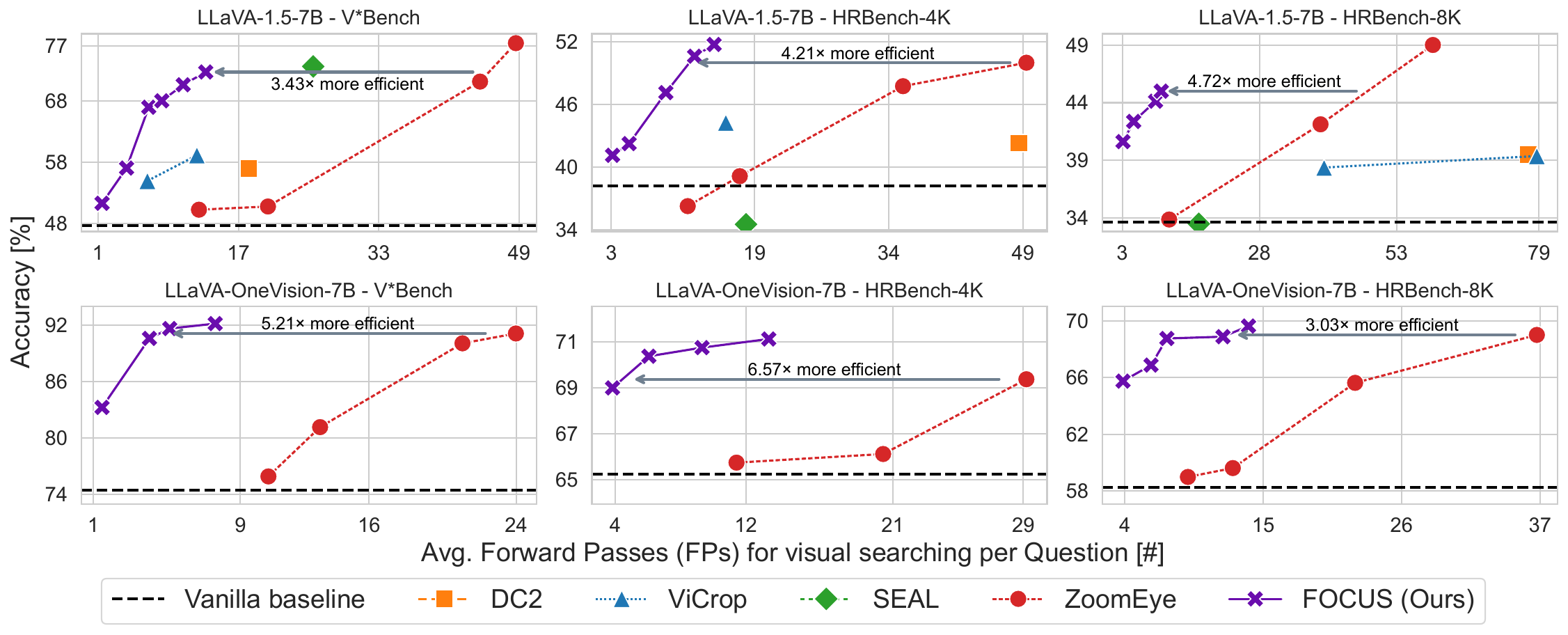}
    \caption{\textbf{\texttt{FOCUS} is at the Pareto front on fine-grained \ac{VQA} benchmarks.} Given the same computation budget, \texttt{FOCUS} (purple crosses) significantly outperforms other visual cropping methods, on three different datasets and for two model architectures. It achieves $3\,\text{--}\,6.5 \times$ higher efficiency than the best-performing baseline ZoomEye. Note that we show only a limited set of data points for each method to ensure a clear visualization. The full results are available in \Cref{appendix:full_numerical_results}.}
    \label{fig:pareto}
\end{figure}

\subsection{Implementation details}
\vspace{-0.3\baselineskip}
\label{subsect: impl_details}

Following prior work \cite{vstar_wu24, dc2_wang2024, zoomeye_shen2024}, we evaluate \texttt{FOCUS} on several fine-grained \ac{VQA} benchmarks: V*Bench \cite{vstar_wu24}, HRBench \cite{dc2_wang2024}, and MME-RealWorld-Lite \cite{mmerealworld_zhang2025}.
We report accuracy on multiple-choice questions as the primary performance metric.
Another critical consideration is the trade-off between performance and efficiency, as inference with \acp{MLLM} can be computationally expensive.
In the case of multiple-choice \ac{VQA}, inference speed is largely determined by the sequence length during the prefill phase \cite{flexgen_sheng, transformer_vaswani}.
Since the sequence length remains relatively consistent across different searches, we quantify efficiency using the number of \acfp{FP} required for the visual search.
We provide details of used hardware and on the calculation of all metrics in \Cref{appendix:hardware_specifications} and \Cref{appendix:metrics}.
We evaluate our method using two types of \acp{MLLM}: LLaVA-1.5-7B \cite{llava_15_liu}, a global-view \ac{MLLM}, and LLaVA-OneVision-7B \cite{llava_ov_li}, a global-local-view \ac{MLLM}.
As described in \Cref{subsect:rel_map}, we utilize representations from multiple layers to compute the object relevance map.
For LLaVA-1.5, we use representations from the $14^{\text{th}}$ to the $32^{\text{nd}}$ layer ($l=14$, $L=32$), and for LLaVA-OneVision from the $14^{\text{th}}$ to the $28^{\text{th}}$ layer ($l=14$, $L=28$).
To evaluate performance under varying computational budgets, we adjust only the number of steps, setting $n_\text{steps} \in \{1, 2, 3, 4, 6, 8\}$.
We describe the configuration of the remaining hyperparameters of \texttt{FOCUS} in \Cref{subsect:hyperpara}\rebuttal{; for an analysis of \texttt{FOCUS}'s hyperparameter sensitivity, see \Cref{subsect: ablation}.
}

For ZoomEye, we vary the number of crops per layer and the cropping depth.
For ViCrop, we report results from its best-performing variants, i.e., \texttt{rel-attn} and \texttt{attn-grad}.
For DC\textsuperscript{2}, we determine the \acp{FP} via the base resolution of the vision encoder.
For SEAL, we evaluate the publicly available pre-trained model without modifications.
Additional implementation details are provided in \Cref{subsect:hyperpara_baseline}.
Note that SEAL and ZoomEye use a different inference scheme on V*Bench compared to the open-ended generation approach of \texttt{FOCUS}.
With this alternative scheme, we observe significantly improved performance with ZoomEye on V*Bench.
Further, we observe a notable gap between our ZoomEye results and those reported in the original paper.
Detailed results are provided in \Cref{subsect:infer_scheme}.

\subsection{Results on fine-grained \ac{VQA} datasets}
\vspace{-0.3\baselineskip}
\label{subsect:results}

We conduct experiments with LLaVA-1.5 on V*Bench, HRBench-4K, and HRBench-8K (see \mbox{\Cref{fig:pareto}}).
Overall, \texttt{FOCUS} outperforms the four other visual cropping methods on a relatively small computational budget of fewer than $17$ \acp{FP}.
\texttt{FOCUS} achieves an accuracy of $72.77\%$  on V*Bench, $51.75\%$ on HRBench-4K, and $45.00\%$ on HRBench-8K.
SEAL achieves a slightly higher accuracy than \texttt{FOCUS} on V*Bench, but at the cost of significantly lower efficiency due to its multi-module design and recursive visual search.
On high-resolution datasets like HRBench, its performance is only on par with the vanilla \ac{MLLM}.
We suspect this decline stems from SEAL’s training data, which is specifically optimized for resolutions below 2K.
ZoomEye achieves the highest accuracy on V*Bench ($77.48\%$), but only with an extremely deep tree search, resulting in substantial computational overhead. 
At our top accuracy of $72.77\%$, \texttt{FOCUS} is $3.43\times$ more efficient than ZoomEye.
On HRBench-4K, our method not only surpasses ZoomEye with a better top accuracy but also uses $4.39\times$ fewer \acp{FP}.
On HRBench-8K, while ZoomEye attains a higher top accuracy of $49.00\%$, \texttt{FOCUS} achieves $44.88\%$ with $ 4.72\times$ greater efficiency.
This performance gap on HRBench-8K is likely due to the limitations of the $24\times24$ object relevance map produced by LLaVA-1.5.
Smaller objects often remain undetected, limiting our model’s ability to improve accuracy—even when increasing $n_{\text{steps}}$ to allocate more~computation.
\begin{table*}[b]
  \caption{\textbf{Results on MME-RealWorld-Lite.} We provide the accuracy for each task. Further, we report the average accuracy and \acp{FP} per subset for efficiency comparison. The dataset includes the domains \textit{OCR (Optical Character Recognition)}, \textit{RS (Remote Sensing)}, \textit{DT (Diagram and Table)}, \textit{MO (Monitoring)}, and \textit{AD~(Autonomous~Driving)}.}
  \label{table:mme_lite}
\footnotesize
    \begin{center}
    \resizebox{\textwidth}{!}{%
    \renewcommand{\arraystretch}{1.2}
    \begin{tabular}{l c c c c c c c c c c c c c c} 
        \toprule
& \multicolumn{7}{c}{\textbf{Perception}} & & \multicolumn{6}{c}{\textbf{Reasoning}} \\
         & \multicolumn{5}{c}{\textbf{Sub-task accuracy} $[\%]$} & \multicolumn{2}{c}{\textbf{Average}} & & \multicolumn{4}{c}{\textbf{Sub-task accuracy} $[\%]$} & \multicolumn{2}{c}{\textbf{Average}} \\
         \textbf{Model} & OCR & RS & DT & MO & AD & \textbf{Acc. $[\%]$} $\uparrow$ & \textbf{FP $[\#]$} $\downarrow$ & & OCR & DT & MO & AD & \textbf{Acc. $[\%]$} $\uparrow$ & \textbf{FP $[\#]$} $\downarrow$ \\
         
        \specialrule{0.1em}{0.1em}{0.1em} 
        \textit{LLaVA-OV-7B} & \cellcolor{gray!10}81.60 & \cellcolor{gray!10}\textbf{52.00} & \cellcolor{gray!10}65.00 & \cellcolor{gray!10}34.48 & \cellcolor{gray!10}43.14 & 52.01 & - & \vline & \cellcolor{gray!10}\textbf{72.00} & \cellcolor{gray!10}40.00 & \cellcolor{gray!10}44.00 & \cellcolor{gray!10}32.35 & 40.93 & - \\
         w/ ZoomEye & \cellcolor{gray!10}81.20 & \cellcolor{gray!10}51.33 & \cellcolor{gray!10}\textbf{74.00} & \cellcolor{gray!10}38.87 & \cellcolor{gray!10}\textbf{51.43} & \textbf{56.29} & 41.60 & \vline & \cellcolor{gray!10}64.00 & \cellcolor{gray!10}49.00 & \cellcolor{gray!10}46.00 & \cellcolor{gray!10}\textbf{35.50} & 43.20 & 45.95 \\
         w/ \texttt{FOCUS} (Ours) & \cellcolor{gray!10}\textbf{83.60} & \cellcolor{gray!10}46.67 & \cellcolor{gray!10}58.00 & \cellcolor{gray!10}\textbf{41.07} & \cellcolor{gray!10}47.14 & 54.15 & \textbf{ 7.71} & \vline & \cellcolor{gray!10}71.00 & \cellcolor{gray!10}\textbf{52.00} & \cellcolor{gray!10}\textbf{51.33} & \cellcolor{gray!10}33.50 & \textbf{44.53} & \textbf{ 8.21} \\
    \bottomrule
    \end{tabular}%
  }
  \end{center}
\end{table*}

With LLaVA-OneVision, we conduct experiments on the previously introduced datasets (see \Cref{fig:pareto}) and additionally evaluate on MME-RealWorld-Lite (see \Cref{table:mme_lite}).
We compare \texttt{FOCUS} only to ZoomEye, as SEAL does not provide any models based on LLaVA-OneVision.
Moreover, neither ViCrop nor DC\textsuperscript{2} supports evaluation with LLaVA-OneVision.
\texttt{FOCUS} significantly benefits from the higher-resolution object relevance map generated based on the local crops.
As a result, our method outperforms ZoomEye both in terms of accuracy and efficiency on the three datasets. 
Although LLaVA-OneVision natively supports resolutions up to 2K, \texttt{FOCUS} can still boost the accuracy on V*Bench (2K) from $74.46\%$ to $92.15\%$.
This can be attributed to our method isolating target objects and reducing irrelevant background regions.
Furthermore, we perform additional evaluation on MME-RealWorld-Lite, see \Cref{table:mme_lite}.
\texttt{FOCUS} outperforms the vanilla baseline on most sub-tasks.
ZoomEye and our method have strengths in different domains.
\texttt{FOCUS} is better for reasoning, while~ZoomEye~is better at perception tasks.
Still, our method is on average $5.47\times$ more efficient than ZoomEye.
We provide the full numerical results in \Cref{appendix:full_numerical_results}.

\textbf{Comparison with ZoomEye}\hspace{1.8mm}
ZoomEye’s strong performance on fine-grained \ac{VQA} is largely driven by its exhaustive tree search, which includes a high number of image regions.
This leads to a substantial number of \acp{FP}, as each region requires three \acp{FP} with different prompts for confidence prediction—making the overall process computationally expensive.
In contrast, \texttt{FOCUS} leverages an object relevance map derived from internal representations to identify target object locations with just a single \ac{FP}—explaining its superior efficiency compared to ZoomEye.
We report additional efficiency metrics, including execution time, FLOPs and memory usage, in \Cref{appendix:efficiency_measurements}.
Crucially, our analysis shows that reducing \acp{FP} not only lowers theoretical computation but also leads to significantly shorter execution times, confirming the practical efficiency of \texttt{FOCUS} compared to ZoomEye.

\subsection{Ablation studies and hyperparameter sensitivity analysis}
\vspace{-0.3\baselineskip}
\label{subsect: ablation}
 
\begin{figure}[h]
\centering
\begin{minipage}{0.56\textwidth}
  \centering
  \footnotesize
\captionof{table}{\textbf{Ablation studies of \texttt{FOCUS}.} We evaluate the influences of design choices of our method based on accuracy and recall. "rel." is short for "relevance". }
\label{tab:ablation}
\resizebox{1\linewidth}{!}{%
\begin{tabular}{lccccc}
\toprule
\multicolumn{3}{c}{\textbf{Ablation}} & \multicolumn{2}{c}{\textbf{V*Bench}} & \textbf{HRBench-4K} \\
 \cmidrule(lr){1-3}\cmidrule(lr){4-5} \cmidrule(lr){6-6}
\multirow{3}{*}{\textbf{Component}}& \makecell{Object \\ rel. map} & \makecell{Proposal \\ ranking} & \makecell{Acc.\\$[\%] \uparrow$}& \makecell{Recall \\$[\%] \uparrow$} &\makecell{Acc.\\$[\%]\uparrow$} \\
\cmidrule{2-3}\cmidrule{4-6}
          & \xmark         & \cmark &     48.68     &   18.37          &    36.13       \\
          & \cmark &  \xmark       &   51.30        &    38.48          &  41.13         \\
\midrule
\textbf{Pseudo-attn.} & \multicolumn{2}{c}{K-K (w/o RoPE)} & 69.10 & 63.47 & 45.63 \\
\midrule
\multirow{2}{*}{\textbf{Layers}} & \multicolumn{2}{c}{$0-14$} & 66.49 & 76.17 & 47.38 \\
                       & \multicolumn{2}{c}{$0-32$} & 71.20 & 75.56 & 49.38 \\
\specialrule{.1em}{.2em}{.2em} 
\multicolumn{3}{l}{\textbf{Original design choice}} & \textbf{72.77} & \textbf{77.49} & \textbf{51.75} \\
\multicolumn{3}{l}{Vanilla baseline} & 47.64 & - & 36.13 \\
\multicolumn{3}{l}{Random guess} & 35.99 & - & 25.00 \\
\bottomrule
\end{tabular}
}

\end{minipage}
\hfill
\begin{minipage}{0.42\textwidth}
  \centering
  \includegraphics[width=0.9\linewidth]{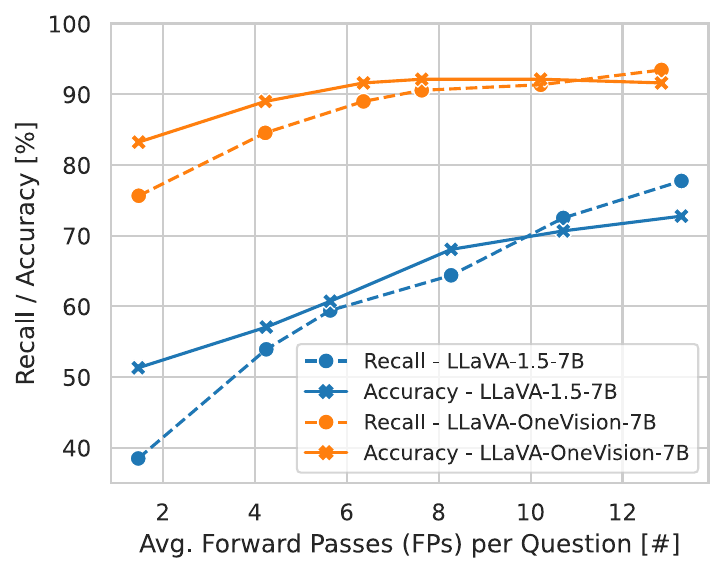}
  \captionof{figure}{\textbf{Ablation studies on the search steps.} We analyze how the number of search steps influences the recall and accuracy on V*Bench. A positive correlation between accuracy and recall can be observed.}
  \label{fig:recall}
\end{minipage}%
\end{figure}

In this subsection, we validate the design choices of \texttt{FOCUS} on V*Bench and HRBench-4K, using LLaVA-1.5 (see \Cref{tab:ablation} and \Cref{fig:recall}).
V*Bench is the only dataset that provides \ac{GT} region annotations, allowing us to report both accuracy and recall.
We calculate recall by checking whether one of the proposed \acp{ROI} overlaps with the \ac{GT} region by at least 50\%.
We use the same hyperparameters across all ablation studies, with the number of steps set to $n_\text{steps} = 8$ unless stated otherwise.
\rebuttal{Furthermore, we analyze the hyperparameter sensitivity of \texttt{FOCUS} on V*Bench.}

\textbf{Component analysis}\hspace{1.8mm}
We first verify the contributions of the object relevance map (see \Cref{subsect:rel_map}) and the \ac{ROI} ranking (see \Cref{subsect:proposal_ranking_roi}).
To assess the impact of the object relevance map, we replace it with a randomly generated map.
This substitution leads to a substantial decrease in accuracy on both datasets—interestingly, however, \texttt{FOCUS} still performs well above random guessing $(35.99\%)$.
This indicates that even without an accurate global understanding of the object’s location, our method can identify likely regions through confidence-based \ac{ROI} selection, demonstrating our ranking mechanism's robustness.
Next, we assess the effect of discarding \ac{ROI} ranking by directly selecting the \ac{ROI} with the highest object relevance score for the final \ac{VQA}.
This yields a recall of $38.48\%$ on V*Bench, with accuracy improvements of $3$ percentage points (pp.) on V*Bench and $5$ pp. on HRBench-4K compared to the vanilla baseline.
These results suggest that object relevance maps, even without post-processing, serve as a strong prior for identifying meaningful visual regions.

\textbf{Value features vs. key features in pseudo-attention}\hspace{1.8mm}
One might argue that key features—central to standard attention—encode richer semantic information \cite{proxyclip_lan} compared to value features and therefore could be used to generate more precise object relevance maps.
However, directly substituting key features for value features in \texttt{FOCUS} results in degraded performance.
This is mainly due to the use of Rotary Positional Embedding (RoPE) \cite{rope_su}  in recent \acp{MLLM} \cite{llava_15_liu, llava_ov_li, qwen_2_5_vl_bai, instruct_blip_dai}, which injects position-dependent rotations into the key features.
As a result, RoPE causes nearby tokens to exhibit artificially high cosine similarity due to positional proximity, rather than semantic alignment \cite{rope_su, feather_endo}.
In this ablation, we remove RoPE from the key features to isolate semantic content before computing object relevance maps.
Despite this, both accuracy and recall remain lower than those achieved using value features (see \Cref{tab:ablation}).
We hypothesize that removing RoPE disrupts the semantic integrity of the key features, as they are trained to operate with positional encoding, resulting in noisier relevance~maps.

\textbf{Later layers vs. earlier layers}\hspace{1.8mm}
Apart from using the default later layers ($14-32$) for LLaVA-1.5, we also experiment with representations from earlier layers ($0-14$) or all layers ($0-32$).
We observe that the representations from later layers yield the best performance across both datasets.
This is consistent with findings from the Logit Lens technique \cite{logits_lens}, which suggests that later layers encode more predictive and semantically coherent representations.

\textbf{Number of steps}\hspace{1.8mm}
In \texttt{FOCUS}, we vary only the number of steps of the \ac{ROI} ranking in our method (see \Cref{subsect: impl_details}).
For this ablation, we report accuracy and recall on V*Bench using LLaVA-1.5 and LLaVA-OneVision.
Across both models, recall increases with more steps, as additional lower-priority \acp{ROI} are explored and more of the image is covered (see \Cref{fig:recall}).
Similarly, accuracy also improves with more steps but begins to plateau beyond a certain point.
This saturation might occur because the model fails to identify the correct region among a considerable number of proposed \acp{ROI} during ranking.
Another reason could be that even when the correct region is provided, the \ac{MLLM} is unable to give the correct answer, due to object size, ambiguity, or other limitations.

\rebuttal{\textbf{Hyperparameter analysis}\hspace{1.8mm}
We investigate \texttt{FOCUS}'s sensitivity on key hyperparameters and find it to be robust across a wide range of parameter choices. In particular, we do not observe any accuracy degradation larger than $4.71$ pp. for LLaVA-1.5 and $2.62$ pp. for LLaVA-OV.
The complete results of this hyperparameter analysis are provided in \Cref{subsect:full_robustness_analysis}.
Please note that this study was conducted post-hoc and not used to optimize the performance of \texttt{FOCUS}.}

\subsection{Additional results}
\vspace{-0.3\baselineskip}
\label{subsect: robustness_analysis}

\rebuttal{In this subsection, we analyze how \texttt{FOCUS} performs on \ac{VQA} questions involving large-size objects. Moreover, we provide results for \texttt{FOCUS} when using the state-of-the-art model Qwen-2.5-VL as the base \ac{MLLM}.}

\rebuttal{\textbf{Results on \ac{VQA} datasets with large objects}\hspace{1.8mm}
While the previously used datasets focus on fine-grained \ac{VQA}, we also evaluate the performance of \texttt{FOCUS} on datasets featuring large-scale objects to assess its robustness across varying object sizes.
We compare \texttt{FOCUS} on the datasets A-OKVQA \cite{aokvqa_schwenk_2022} and GQA \cite{gqa_hudson_2019} using LLaVA-1.5 and LLaVA-OV with the vanilla model and ViCrop as the latter one is the only benchmark method providing respective results (see \Cref{tab:large_scale_datasets}). 
Overall, \texttt{FOCUS} demonstrates strong robustness on datasets with large objects, maintaining competitive performance compared to the baseline models.
}

\rebuttal{\textbf{Results with Qwen-2.5-VL}\hspace{1.8mm}
Qwen-2.5-VL \cite{qwen_2_5_vl_bai} processes high-resolution images with native resolution, thereby preserving spatial details more effectively.
We evaluate \texttt{FOCUS} with Qwen-2.5-VL-7B (see \Cref{tab:results_qwen_25_vl}) and find state-of-the-art accuracy on HRBench-4K and HRBench-8K. This confirms the compatibility of \texttt{FOCUS} with different \ac{MLLM} architectures.}

\begin{figure}[h]
\centering
\begin{minipage}[t]{0.48\textwidth}
  \centering
  \footnotesize
\captionof{table}{\textbf{Results on \ac{VQA} datasets with large objects.} We find only minor performance degradation of \texttt{FOCUS} w.r.t.~the base model.}
\label{tab:large_scale_datasets}
\resizebox{1\linewidth}{!}{%
    \begin{tabular}{l c c c c}
    \toprule
    & \multicolumn{2}{c}{\textbf{A-OKVQA}} & \multicolumn{2}{c}{\textbf{GQA}}\\
    \textbf{Model} & Acc. $[\%]$ & $\Delta$ & Acc. $[\%]$ & $\Delta$ \\
    \midrule
    LLaVA-1.5 & 77.99 & - & 61.97 & - \\
    w/ ViCrop & 60.66 & -17.33 & 60.98 & -0.99 \\
    w/ \texttt{FOCUS} & 74.76 & -3.23 & 60.34&-1.63\\
    \midrule
    LLaVA-OV & 91.44 & - &62.01 & -\\
    w/ \texttt{FOCUS} & 91.00 & -0.44 & 51.02&-10.99\\
    \bottomrule
    \end{tabular}
}
\end{minipage}
\hfill
\begin{minipage}[t]{0.48\textwidth}
  \centering
  \footnotesize
\captionof{table}{\textbf{Results of \texttt{FOCUS} with Qwen-2.5-VL.} \texttt{FOCUS} significantly boosts the performance of the base model, consistent with previous results for LLaVA-1.5 and LLaVA-OV.}
\label{tab:results_qwen_25_vl}
\resizebox{1\linewidth}{!}{%
    \begin{tabular}{l c c c}
    \toprule
    \textbf{Model} & \textbf{\makecell{V*Bench \\ $[\%]$}} & \textbf{\makecell{HRBench-4K \\ $[\%]$}} & \textbf{\makecell{HRBench-8K \\ $[\%]$}} \\
    \midrule
    Qwen-2.5-VL & 79.06 & 71.62 & 68.62 \\
    w/ \texttt{FOCUS} & \textbf{90.58} & \textbf{79.25} & \textbf{76.25}\\
    \bottomrule
    \end{tabular}
}
\end{minipage}%
\end{figure}

\subsection{Qualitative examples}
\begin{figure}[h]
    \centering
    \includegraphics[clip, trim=0cm 2.0cm 8.0cm 0.3cm, width=1\textwidth]{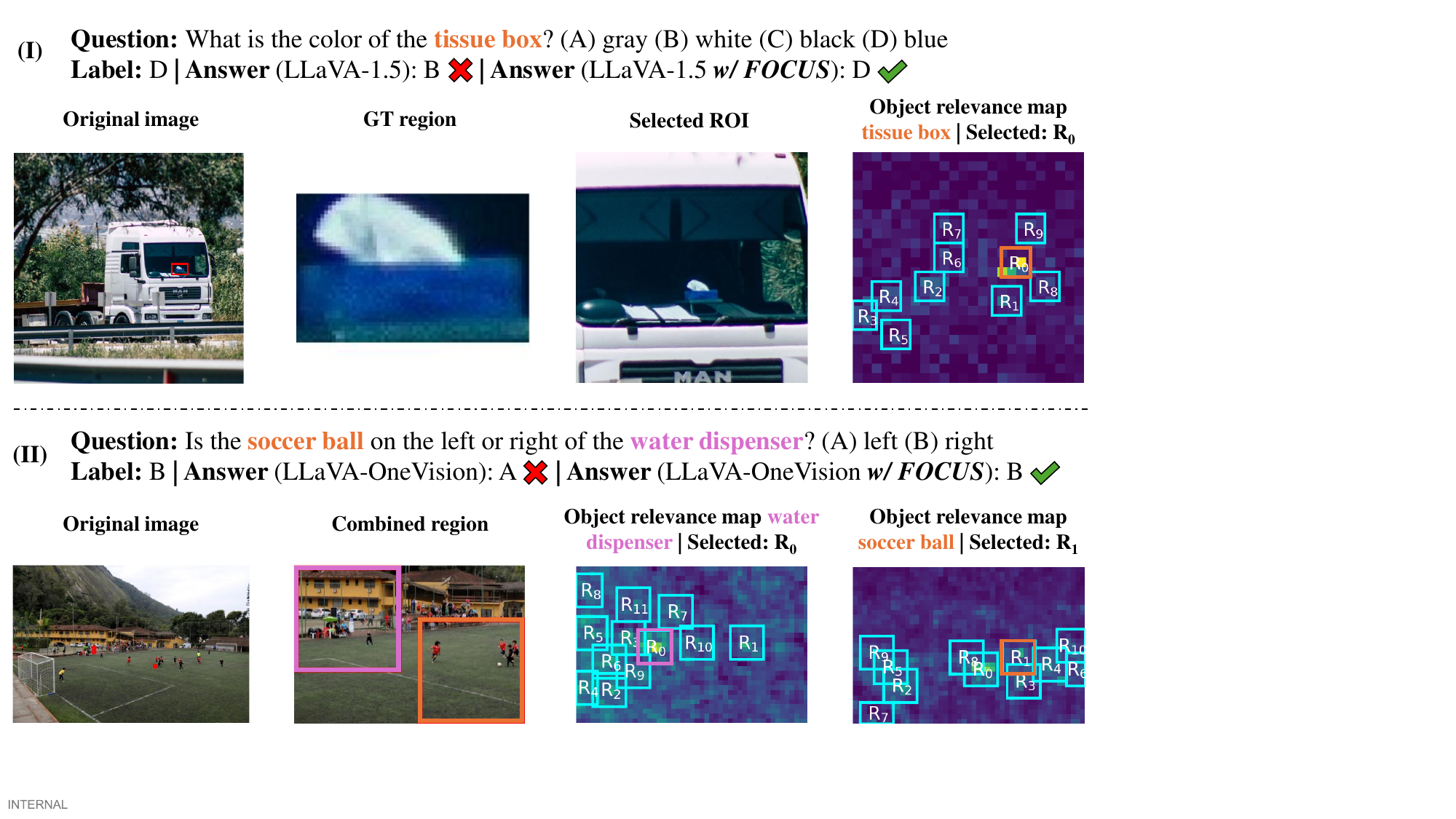}
    \caption{\textbf{Qualitative examples of \texttt{FOCUS}.} We provide some exemplary inferences with our method for single-target tasks with LLaVA-1.5 \textbf{(I)} and multiple-target tasks with LLaVA-OneVision \textbf{(II)}. The \acf{GT} locations are highlighted in red in the original image. Further, we show the detected image regions and their locations in the object relevance map. Note that the object relevance maps corresponds to the original images.}
    \label{fig:qualitative_examples}
\end{figure}

\vspace{-0.3\baselineskip}
\label{subsect:qualitative_examples}
We provide two qualitative examples that show how \texttt{FOCUS} improves performance of LLaVA-1.5 for single-target tasks and LLaVA-OneVision for multiple-target tasks \rebuttal{(see \Cref{fig:qualitative_examples})}.
In both examples, the accurate visual crops generated by \texttt{FOCUS} enable the respective \ac{MLLM} to answer the question correctly.
The detected location of the target objects is highlighted in the object relevance map.
One can see that LLaVA-OneVision provides a higher-resolution and cleaner object relevance map.
We provide more examples in \Cref{sect:qualitative_examples}.

\subsection{Limitations}
\vspace{-0.3\baselineskip}
\label{subsect:limitations}
One limitation of our method is its reliance on the resolution of the object relevance map. When the input image is high-resolution (e.g., 8K), but the internal representations of the \ac{MLLM} can only produce a low-resolution relevance map, accurately localizing fine-grained objects becomes difficult.
This limitation partly explains the reduced effectiveness of our method with LLaVA-1.5 on HRBench-8K.
A potential solution is to construct the object relevance map in a sliding-window manner over the image, allowing finer spatial resolution.
\rebuttal{Moreover, \texttt{FOCUS} inherits the typically limited understanding of spatial relationships \cite{limit_tong24, limit2_chen24} from the base \ac{MLLM}, as it is a training-free method. Thus, \texttt{FOCUS} struggles with spatial concepts such as "on the left/right of the image".} We leave these shortcomings for future work.

\section{Conclusion}
\vspace{-0.5\baselineskip}
\label{sec:conclusion}

In this work, we proposed \texttt{FOCUS}, an efficient, training-free visual cropping method for fine-grained \ac{VQA} tasks, where identifying small objects is essential.
Our method constructs an object relevance map from cached token representations to localize image regions relevant to the question, enabling detail-focused \ac{VQA} inference.
\texttt{FOCUS} achieves performance on par with or better than existing methods across multiple fine-grained \ac{VQA} benchmarks, while requiring significantly fewer computational overhead.
These results highlight the potential of training-free, high-resolution \ac{VQA} systems that are both effective and computationally efficient. \rebuttal{Moreover, the central idea of \texttt{FOCUS}---harnessing the hidden spatial capabilities of MLLMs via an inference-time method---holds significant promise for spatial reasoning tasks well beyond VQA.}

\newpage
\section*{Acknowledgment}
The authors gratefully acknowledge financial support by the German Federal Ministry for Economic Affairs and Energy under the grant numbers 19A24004U and 19A24004I as a part of the \textit{Safe AI Engineering} consortium.

DISCLAIMER: The results, opinions and conclusions expressed in this publication are not necessarily those of Volkswagen Aktiengesellschaft.

\bibliographystyle{plainnat}
\bibliography{references}
\newpage

\appendix
\normalsize

\section{Implementation details}
\label{appendix:implementation_details}

In this section, we provide additional technical details of our work.
We begin by describing the hardware setup and report the total GPU hours required to reproduce our main results in \Cref{appendix:hardware_specifications}.
Next, we explain in detail how the reported metrics are calculated, see \Cref{appendix:metrics}.
We then provide the hyperparameters used for \texttt{FOCUS}
and the baseline visual cropping methods in \Cref{subsect:hyperpara} and \Cref{subsect:hyperpara_baseline}, respectively.
Finally, we present further details on the processing scheme of \texttt{FOCUS} in~\Cref{subsct:final_vqa_infer}.

\subsection{Hardware specifications}
\label{appendix:hardware_specifications}

We run all experiments presented in \Cref{sect:experiments} and \Cref{subsect:further_results} on identical hardware, namely compute instances equipped with an \textit{NVIDIA A100 80GB} GPU, an \textit{AMD EPYC 7V13} CPU, and \textit{220 GB} of RAM.
\texttt{FOCUS}'s software environment includes \textit{CUDA 12.2}, \textit{PyTorch 2.6.0}, and the HuggingFace \textit{transformers} library in version \textit{4.46.0}.

We also provide an estimation of the GPU hours needed to reproduce our results in the entire project.
In total, our reported results require approximately 132 GPU hours on the hardware configuration described above (see~\Cref{tab:gpu_hours}).
Additionally, we report the GPU hours required to run the experiments that achieve the best-performing variants for each method on LLaVA-1.5 and LLaVA-OneVision.

\begin{table*}[hbtp]
    \centering
    \caption{\textbf{Estimated GPU hours.}}
    \small
    \label{tab:gpu_hours}
    \renewcommand{\arraystretch}{1.2}
    \begin{tabular}{lccc}
         \toprule
         \textbf{Model} & \textbf{\makecell{Best-perf. variant \\ w/ LLaVA-1.5}} & \textbf{\makecell{Best-perf. variant \\ w/ LLaVA-OV}} & \textbf{All variants}\\
         \midrule
         \textbf{SEAL} & $6$ \footnotemark & -- & $6$\\
         \midrule
         \textbf{ViCrop} & $3$ & -- & $6$ \\
         \textbf{ZoomEye} & $5$ & $30$ & $80$\\ 
         \textbf{\texttt{FOCUS} (Ours)} & $2$ & $9$ & $40$ \\
         \midrule
         \textbf{Sum} & $16$ & $39$ & $132$ \\
         \bottomrule
    \end{tabular}
\end{table*}

\footnotetext{SEAL uses customized \acp{MLLM} based on LLaVA-7B.}

\subsection{Metrics}
\label{appendix:metrics}

In this subsection, we provide a detailed description of our performance and efficiency metrics.

\textbf{Performance metrics}\hspace{1.8mm}
The main performance metric is the accuracy of visual cropping methods, which is typically reported as a percentage. 
We follow the standard evaluation protocol and do not apply any post-processing to the answers generated by the visual cropping methods using the \ac{MLLM}.  
This means that if the ground-truth label is \texttt{"A"} and the model outputs \texttt{"The answer is A"} or \texttt{"(A)"}, we do not extract or normalize the answer to match the label.  
As a result, such responses are counted as incorrect. However, this type of mismatch is of minor importance in our experiments: across all models and three datasets (V*Bench, HRBench-4K, HRBench-8K), we did not observe any irregular or non-standard response formats.
We compute the average accuracy as the unweighted mean over all $N$ samples in a dataset, i.e., as $\sum_{i=1}^{N} (\hat{y}_i = y_i)/N$, where $\hat{y}_i$ is the predicted answer and $y_i$ is the ground-truth label for the $i$-th sample.

\textbf{Efficiency metrics}\hspace{1.8mm}
An important metric is the number of \acfp{FP} required for the visual search. Note that we exclude the \acp{FP} needed for the final \acp{VQA} prediction, as different inference schemes can generate different forward passes, see \Cref{subsect:infer_scheme}. We compute this by tracking how often the \texttt{generate}, \texttt{forward}, or \texttt{\_\_call\_\_} methods of the respective \ac{MLLM} are invoked per question. For methods that use multiple \acp{MLLM} (e.g., SEAL), we report the total number of \acp{FP} of all \acp{MLLM}.

Another key metric we report is the efficiency improvement of \texttt{FOCUS} relative to the baseline methods.
Among all evaluated approaches, \texttt{FOCUS} demonstrates the highest efficiency.
To ensure a fair comparison, we consider the top accuracy achieved by \texttt{FOCUS} and the best-performing baseline, and select the lower of the two as the reference accuracy.
For both methods, we determine the number of \acfp{FP} required to reach this reference accuracy—either by taking the exact value or interpolating between data points to estimate the \acp{FP}.
This yields $\text{FP}_{\text{ours}}$ for \texttt{FOCUS} and $\text{FP}_{\text{ref}}$ for the best-performing baseline.
The relative efficiency improvement is then computed as $\text{FP}_{\text{ref}} / \text{FP}_{\text{ours}}$.

Furthermore, we report two additional metrics to provide a practical comparison of efficiency, i.e., execution time and peak GPU memory usage (see \Cref{tab:add_efficiency}).
Execution time is recorded per sample, and we report the average time per question across a dataset.
Peak memory is measured using \texttt{torch.cuda.max\_memory\_allocated()} and converted to GB by dividing the result by $1024^3$.

\subsection{Hyperparameters of \texttt{FOCUS}}
\label{subsect:hyperpara}
This subsection outlines the hyperparameters used in our method, \texttt{FOCUS}. These parameters are applied consistently across all experiments and correspond to the results reported in \Cref{sect:experiments}.
As noted in the main paper, the only parameter we vary is the number of steps, $n_{\text{steps}}$.
Most other hyperparameters remain fixed across experiments for both LLaVA-1.5 and LLaVA-OneVision.
The exceptions are three parameters: $k$ and $s_{\text{dist}}$ (specific to LLaVA-1.5), and $s_{\text{max}}$ (specific to LLaVA-OneVision), as detailed in \Cref{tab:hyperparameter_focus}.

For LLaVA-1.5, we use a smaller value of $k$ when $n_{\text{steps}}$ is low to reduce the number of proposed \acp{ROI}.
Additionally, we increase $s_{\text{dist}}$ on HRBench to ensure a broader spatial distribution of the \acp{ROI} across the image.
In contrast, for LLaVA-OneVision, we set a larger $k$ due to its higher-resolution object relevance map.
Moreover, we set $s_{\text{max}}=9$ for V*Bench and $s_{\text{max}}=5$ for the other datasets, as V*Bench contains lower-resolution~(2K)~images.

The hyperparameters used in \texttt{FOCUS} are generally robust and transferable across a wide range of use cases. For users applying our method to new datasets, we recommend adjusting $s_{\text{max}}$, which determines the maximum size of each proposed \acp{ROI}, based on both the resolution of the input images and the spatial resolution of the object relevance map. In particular, $s_{\text{max}}$ should be chosen so that the corresponding region in the original image spans approximately $1$–$2\times$ the base resolution of the vision encoder. For example, if the object relevance map has a spatial resolution of $60\times30$, and the input image resolution is $7680\times3840$, then each grid element corresponds to an area of $128\times128$. Setting $s_{\text{max}}=5$ yields a maximum crop size of $640\times640$, which falls within the recommended range of $384\times384 - 768\times768$ for the SigLIP encoder. Additionally, if one considers migrating our method to another \ac{MLLM}, we recommend selecting the last $25\% - 60\%$ of the layers.

\begin{table*}[hbtp]
    \centering
    \caption{\textbf{Hyperparameters of \texttt{FOCUS}.}}
    \label{tab:hyperparameter_focus}
    \renewcommand{\arraystretch}{1.35}
    \resizebox{\textwidth}{!}{
    \begin{tabular}{cp{4.5cm}cc}
         \toprule
         \textbf{\makecell{Hyper- \\ parameter}} & \textbf{Description} & \textbf{LLaVA-1.5} & \textbf{LLaVA-OneVision} \\
         \midrule
         $k$ & Number of anchor points & $k=\begin{cases} 15 & \text{if $n_{\text{steps}} < 4$} \\ 30 & \text{otherwise}\end{cases}$ & $k=30$ \\
         $s_{\text{min}}$ & Minimum size of each \ac{ROI} & $s_{\text{min}}=3$ & $s_{\text{min}}=3$ \\
         $s_{\text{max}}$ & Maximum size of each \ac{ROI} & $s_{\text{max}}=5$ & $s_{\text{max}}=\begin{cases} 9 & \text{for V*Bench} \\ 5 & \text{otherwise}\end{cases}$ \\
         $s_{\text{dist}}$ & Minimum Euclidean distance between anchor points & $s_{\text{dist}}=\begin{cases} 2 & \text{for V*Bench} \\ 3 & \text{otherwise}\end{cases}$ & $s_{\text{dist}}=2$ \\
         $l$ & Start layer of the used \ac{MLLM}-internal representations & $l=14$ & $l=21$  \\
         $L$ & End layer of the used \ac{MLLM}-internal representations & $L=32$ & $L=28$ \\
         $t_{\text{type2}}$ & Threshold for inclusion of \acp{ROI} for type-2 questions & $t_{\text{type2}}=0.6$ & $t_{\text{type2}}=0.5$  \\
         $t_{\text{obj\_dist}}$ & Threshold for merging \acp{ROI} of nearby objects
         (see \Cref{subsct:final_vqa_infer}) & $t_{\text{obj\_dist}} = 1200$ & -- \\

         \bottomrule
    \end{tabular}%
    }
\end{table*}

\subsection{Hyperparameters of recent visual cropping methods}
\label{subsect:hyperpara_baseline}

\begin{figure}[hbtp]
    \centering
    \includegraphics[clip, trim=0cm 4.5cm 1cm 5.2cm, width=0.85\linewidth]{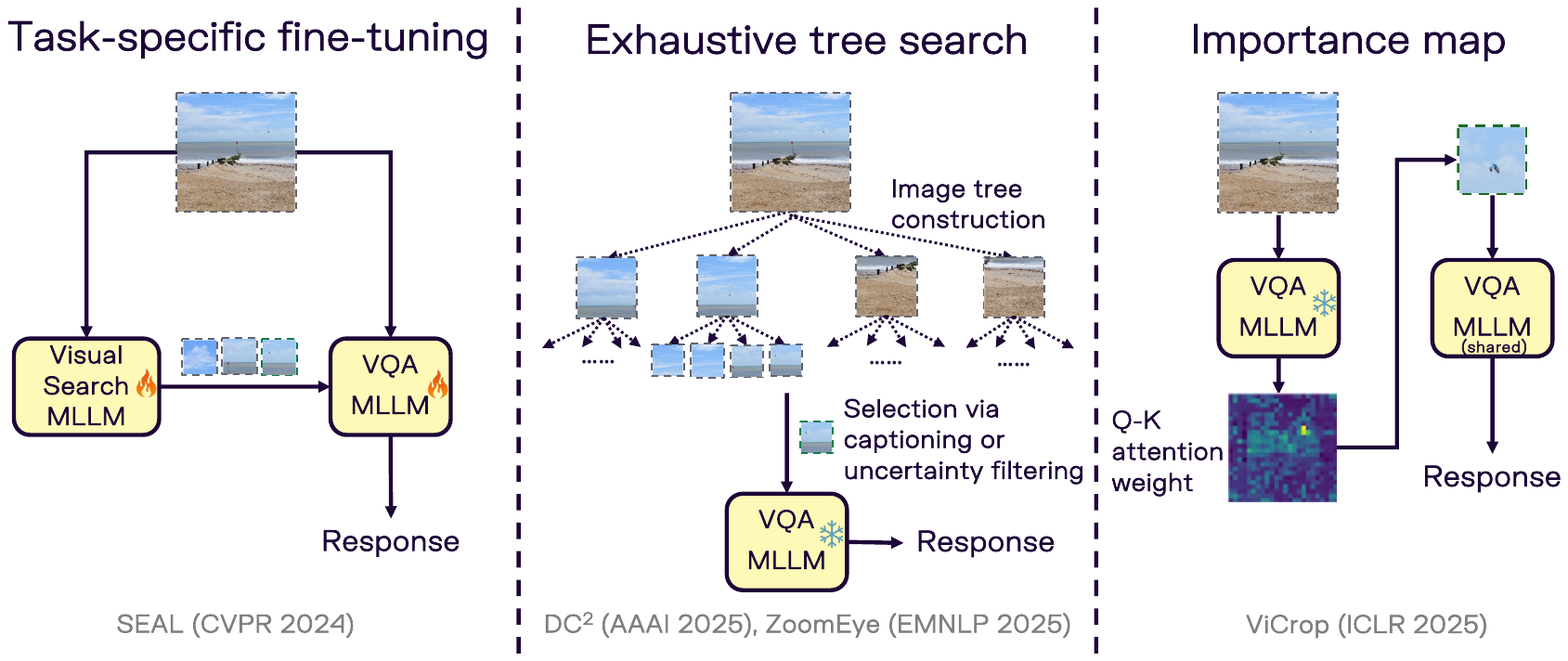}
    \caption{\textbf{Comparison of recent visual cropping methods.} We categorize the baseline methods SEAL, DC\textsuperscript{2}, ZoomEye, and ViCrop based on whether they: (1) require task-specific fine-tuning, (2) employ exhaustive tree search to identify relevant image regions, or (3) compute importance maps using attention matrices.}
    \label{fig:comparison_baseline_methods}
\end{figure}

This subsection outlines the hyperparameters used for the baseline methods, i.e., for DC\textsuperscript{2}, SEAL, ViCrop, and~ZoomEye.
Further, we provide a visual comparison of these baseline methods in \Cref{fig:comparison_baseline_methods}.

For DC\textsuperscript{2}, the full evaluation code is not publicly available, and we were unable to reproduce their results using the provided demo code.
Therefore, we report the performance metrics as stated in their paper and estimate the number of \acfp{FP} based on the available demo.
Specifically, we follow the procedure described in the paper: splitting the image into patches using the base resolution of the vision encoder (i.e., $336\times336$ for LLaVA-1.5) and merging patches via hierarchical clustering to improve efficiency.
Although the \acp{FP} for DC\textsuperscript{2} are estimated and may carry a high margin of error, the method remains less efficient than other baselines.
This is further exacerbated by its region-wise captioning step, which makes it more computationally intensive than other baselines, even when the number of \acp{FP} is comparable.
Furthermore, DC\textsuperscript{2} consistently underperforms compared to the other methods across all three datasets—V*Bench, HRBench-4K, and HRBench-8K—a trend also noted in ZoomEye’s evaluation.

For SEAL, we use the hyperparameters described in their paper and the default configuration provided in their code.
Specifically, we set the minimum search size to $224$ and the minimum search scale to $4$. 
For the visual search, we use a confidence lower bound of $0.3$ and a confidence upper bound of $0.5$.
Regarding the target cue, we set the threshold to $6$, the decay factor to $0.7$, and the minimum threshold to $4$.
Note that these parameters were originally configured for the V*Bench dataset, and we did not adjust them for the other datasets.

For ViCrop, we select only the two best-performing variants from their work: \texttt{att-grad-high} and \texttt{rel-att-high}.
Notably, both methods employ the high-resolution processing scheme \texttt{high}, which divides high-resolution images into a grid of $1K$ sub-images and computes importance maps for each sub-image individually.
As a result, the computational overhead increases significantly with higher input resolutions.

For ZoomEye, we report more results per dataset–model combination than those presented in the original paper to offer a more comprehensive view of its efficiency–accuracy trade-offs. Specifically, we vary two key parameters: the number of sub-regions into which each region is split ($2$ or the default $4$ crops), and the depth of the search tree ($1$, $2$, and the default $5$). All other hyperparameters are as specified in the ZoomEye paper.

\subsection{Additional implementation details of \texttt{FOCUS}}
\label{subsct:final_vqa_infer}
We provide additional implementation details of \texttt{FOCUS} to ensure reproducibility, including how to construct the object relevance maps and how to perform the final \ac{VQA} prediction.

\begin{figure}[!t]
  \centering
    \includegraphics[clip, trim=2cm 4.5cm 5cm 2cm, width=0.85\textwidth]{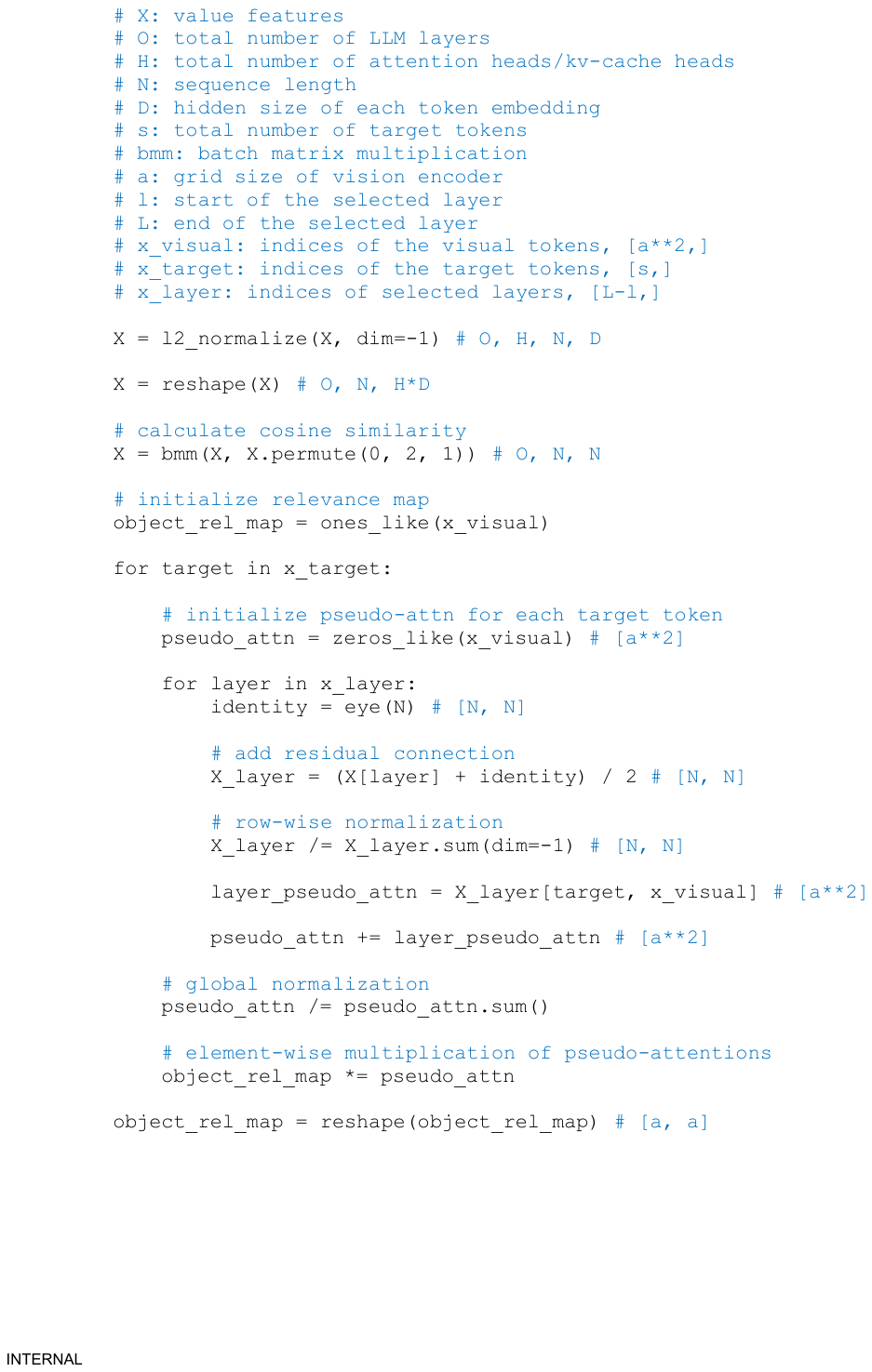}
    \caption{\textbf{PyTorch-style pseudocode for constructing object relevance maps in \texttt{FOCUS}. }}
    \label{fig:core_pseudo_code}
\end{figure}

\textbf{Constructing object relevance maps}\hspace{1.8mm}
We provide PyTorch-style pseudocode in \Cref{fig:core_pseudo_code}. For a detailed motivation and description of this method, see \Cref{sect:method}.

\textbf{Verification of object existence in \ac{ROI}}\hspace{1.8mm}
Inspired by ZoomEye \cite{zoomeye_shen2024}, we use a yes/no prompt \texttt{"Is there a \{target object\} in the image?"} to verify whether the target object exists within a \ac{ROI}. We compute a confidence score $c_{\text{Yes}}$ based on the softmax-normalized logits $l_{\text{Yes}}, l_{\text{No}}$ \mbox{corresponding} to the responses \texttt{"Yes"} and \texttt{"No"}. Specifically, $c_{\text{Yes}}$ is defined as: \begin{equation} 
c_{\text{Yes}} = 2 \cdot (\text{softmax}([l_{\text{Yes}}, l_{\text{No}}])_\text{Yes} - 0.5),\;\;\; c_{\text{Yes}} \in [-1, 1]\;.
\end{equation}

\textbf{Final VQA prediction}\hspace{1.8mm}
\rebuttal{
As explained in \Cref{subsect:proposal_ranking_roi}, we differentiate between type-1 and type-2 questions.
Type-1 questions involve single-object instances, e.g., "\texttt{What is the color of the car?}" or "\texttt{What is the relative position of the ball to the bench?}".
{Type-2} questions concern multiple instances of an object type, e.g.,~"\texttt{How many bikes are in the image?}".}
We follow the inference strategy introduced in ZoomEye \cite{zoomeye_shen2024} when passing the selected \acp{ROI} to the \acp{MLLM}. In this appendix, we elaborate on the strategy used for global-view \acp{MLLM}, which typically accept only a single low-resolution image~input.

For type-1 questions, we select the \ac{ROI} with the highest confidence score for each target object. If the question involves multiple target objects, we compute the combined image region covering all targets and use this as the final VQA input. However, this approach may fail when the target objects are far apart, as the resulting combined region can become excessively large and include irrelevant background.
To address this, a fallback strategy is taken. If the Euclidean distance between any pair of target objects exceeds a threshold $t_{\text{obj\_dist}}$ (e.g., $1200$ pixels), we instead resize each individual \ac{ROI} to a lower resolution and paste them onto an empty canvas based on their relative positions. This avoids including unnecessary background while maintaining spatial relationships. The resulting composite image is then used for the final VQA prediction. For type-2 questions, we first select all \acp{ROI} whose confidence scores exceed a predefined threshold. Overlapping regions are then merged. The merged \acp{ROI} are subsequently placed on a canvas using the pasting strategy described above.

In the case of global-local-view \acp{MLLM}, this workaround is unnecessary, as these models natively support multi-image reasoning without requiring canvas composition. Fine-grained details in local regions can be preserved by supplying them as separate image inputs. We leverage the model’s text–image interleaved capabilities \cite{flamingo_alayrac, llava_next_li, llava_ov_li} by providing a global view - with all \acp{ROI} visually highlighted, alongside the individual high-confidence local~crops.

\textbf{Proposal ranking of  \acp{ROI} for fine-grained \ac{VQA}}\hspace{1.8mm}
\rebuttal{
Further, we present pseudocode in \Cref{algo} that outlines the creation and ranking of \acp{ROI}
to identify the most relevant \ac{ROI} for the final inference, as described in \Cref{subsect:proposal_ranking_roi}.
}

\begin{algorithm}
\renewcommand\thealgorithm{}
\caption{Pseudocode for proposal ranking of ROIs for fine-grained VQA}
\label{algo}
\begin{algorithmic}[1]
    \Require Object relevance map $A$, number of anchor points $k$, number of steps $n_{\text{steps}}$, ROI parameters: $s_{\text{min}}$, $s_{\text{max}}$, $s_{\text{dist}}$; threshold for NMS: $\text{NMS}_{\text{threshold}}$; model MLLM\vspace{1em}
    \State anchor\_points $\leftarrow$ \texttt{Extract-Top-K-Anchors}($A, k$)
    \State initial\_rois $\leftarrow$ \texttt{Generate-Symmetric-ROIs}(anchor\_points, $s_{\text{min}}$)
    \State expanded\_rois $\leftarrow$ \texttt{Expand-ROIs}(initial\_rois, $A, s_{\text{max}}, s_{\text{dist}}$)
    \State filtered\_rois $\leftarrow$ \texttt{Apply-NMS}(expanded\_rois, $\text{NMS}_{\text{threshold}}$)
    \State ranked\_rois $\leftarrow$ \texttt{Rank-ROIs-By-Confidence}(filtered\_rois, $n_{\text{steps}}$)
    \State answer $\leftarrow$ \texttt{Final-Inference}(MLLM, ranked\_rois)\vspace{1em}
    \Ensure MLLM answer based on top-ranked ROI
\end{algorithmic}
\end{algorithm}

\section{Dataset statistics}
This section provides a summary of the datasets used in our experiments.
For each dataset, we report the average image resolution, the number of images, and the number of question–answer pairs in~\Cref{tab:dataset_overview}.

\begin{table*}[hbtp]
    \centering
    \caption{\textbf{Statistics of the employed VQA datasets}}
    \resizebox{0.8\textwidth}{!}{
    \begin{tabular}{lrrrr}
        \toprule
        \textbf{Attribute} & \textbf{V*Bench} & \textbf{HRBench-4K} & \textbf{HRBench-8K} & \textbf{MME-RealWorld-Lite} \\
        \midrule
         Avg. width & 2,246 & 4,024 & 7,431 & 2,836 \\
         Avg. height & 1,582 & 3,503 & 5,358 & 1,566 \\
         Images $[\#]$ & 191 & 200 & 200 & 1,543 \\
         QA-pairs $[\#]$ & 191 & 800 & 800 & 1,919 \\
         \bottomrule
    \end{tabular}%
    }
    \label{tab:dataset_overview}
\end{table*}

V*Bench \cite{vstar_wu24} comprises images with an average resolution of 2K, specifically $\num{2256} \times \num{1582}$ pixels.
The dataset includes two types of tasks: \textit{direct attribute} and \textit{spatial relation}.
The \textit{direct attribute} task involves identifying visual properties of a single object (e.g., color), making them single-target tasks.
In contrast, the \textit{spatial relation} task requires predicting the spatial relationship between two objects, thus constituting multiple-target tasks.
In total, the dataset contains $191$ images, each paired with a single question, resulting in $191$ question-answer~(QA)~pairs.

HRBench \cite{dc2_wang2024} comprises two sub-datasets with average image resolutions of 4K and 8K, respectively.
Both sub-datasets include two task types: \textit{FSP} (Fine-grained Single-instance Perception) and \textit{FCP} (Fine-grained Cross-instance Perception).
\textit{FSP} questions are single-target and focus on identifying fine-grained attributes of individual objects, whereas \textit{FCP} questions are multiple-target and involve reasoning about spatial relationships between target objects.
HRBench-8K contains full-resolution images with an average size of $\num{7431} \times \num{5358}$ pixels, while HRBench-4K provides cropped versions of these images, on average with $\num{4024} \times \num{3503}$ pixels.
Each dataset contains one question per image.
To improve evaluation robustness, HRBench permutes the positions of the answer options, yielding a total of 800 question–answer pairs across 200 images.

MME-RealWorld-Lite \cite{mmerealworld_zhang2025} is a real-world, fine-grained \ac{VQA} dataset with an average image resolution of $\num{2836} \times \num{1566}$ pixels.
It includes a diverse set of subtasks designed to evaluate the perception and reasoning capabilities of \acp{MLLM} across various domains, such as Autonomous Driving and OCR.
The dataset is divided into two subsets: \textit{Perception} and \textit{Reasoning}.
Each subset contains questions from multiple domains, including \textit{OCR} (Optical Character Recognition), \textit{RS} (Remote Sensing), \textit{DT} (Diagram and Table), \textit{MO} (Monitoring), and \textit{AD} (Autonomous Driving), resulting in a total of nine tasks, as shown in \Cref{tab:dataset_mme}.
It comprises $\num{1543}$ images, with some images associated with multiple questions, leading to a total of $\num{1919}$ questions.
The number of QA pairs per task is detailed in \Cref{tab:dataset_mme}.

\begin{table*}[hbtp]
    \centering
    \caption{\textbf{Number of question-answer (QA) pairs in MME-Realworld-Lite per domain}}
    \label{tab:dataset_mme}
    \renewcommand{\arraystretch}{1.2}
    \resizebox{0.7\textwidth}{!}{
    \begin{tabular}{lccccccccc}
        \toprule
        & \multicolumn{5}{c}{\textbf{Perception}} & \multicolumn{4}{c}{\textbf{Reasoning}} \\
        & OCR & RS & DT & MO & AD & OCR & DT & MO & AD \\
        \midrule
         QA-pairs $[\#]$ & 250 & 150 & 100 & 319 & 350 & 100 & 100 & 150 & 400  \\
         \bottomrule
    \end{tabular}
    }
\end{table*}

\section{Additional results}
\rebuttal{
We include full numerical results in \Cref{appendix:full_numerical_results}, results on open-ended \ac{VQA} in \Cref{appendix:open_VQA}, and an analysis of the hyperparameter influence on \texttt{FOCUS} in \Cref{subsect:full_robustness_analysis}.
Further, we provide additional efficiency metrics in \Cref{appendix:efficiency_measurements} and a comparison of different inference schemes as well as performance discrepancies between reported and reproduced results in \Cref{subsect:infer_scheme}.
}
\label{subsect:further_results}

\subsection{Full results}
\label{appendix:full_numerical_results}
We present the full results of ZoomEye \cite{zoomeye_shen2024}, ViCrop \cite{vicrop_zhang2025}, DC\textsuperscript{2} \cite{dc2_wang2024}, SEAL \cite{vstar_wu24}, and our method \texttt{FOCUS} on V*Bench, HRBench-4K, and HRBench-8K in \Cref{table:main results_full}.
For MME-RealWorld-Lite \cite{mmerealworld_zhang2025}, we compare ZoomEye, the vanilla baseline, and \texttt{FOCUS}, see \Cref{table:mme-real-world-full}.

As described in \Cref{subsect: impl_details}, we run experiments with 
$n_\text{steps} \in \{1,2,3,4,6,8\}$. To better leverage the reasoning capabilities of \acp{MLLM} under higher computational budgets, we allow an \texttt{overrun} mode:
if the logit $l_\text{Yes}$ is lower than $l_\text{No}$—i.e., the \ac{MLLM} responds \texttt{"No"} to the existence prompt for all of the top-$n_\text{steps}$ \acp{ROI}—the model continues evaluating additional \acp{ROI} until it receives a \texttt{"Yes"} response. This mechanism improves the efficiency trade-off in many cases. For $n_\text{steps} \in \{1,2\}$, we report results both with and without the \texttt{overrun} mechanism to provide a complete comparison in the low-computation setting.

In general, accuracy improves with increased computation budget for methods that support a configurable computation budget—such as ZoomEye and \texttt{FOCUS}.
ZoomEye exhibits \textit{exponential} scaling in the number of evaluated regions due to its hierarchical tree structure, leading to a significantly higher number of \acfp{FP} when targeting high accuracy.
In contrast, \texttt{FOCUS} constructs an object-aware relevance map and directly retrieves the most relevant regions, resulting in \textit{linear} scaling with respect to the number of \acp{FP}.
\newcommand{\rowheight}{3.5ex}
\begin{table*}[hbtp]
\footnotesize
  \begin{center}
    \caption{\textbf{Full results of different models on fine-grained VQA benchmarks.} V*Bench comprises two tasks, namely direct attribute (Attr) and spatial relationship (Spatial). Similarly, HRBench consists of two tasks, i.e.,~Fine-grained Single-instance Perception (FSP) and Fine-grained Cross-instance Perception (FCP). The highest accuracy for each method-model combination is highlighted in bold. As DC\textsuperscript{2} does not provide the complete evaluation code, we report the accuracy from the original paper and estimate the number of \acp{FP} following the procedure described in \Cref{subsect:hyperpara_baseline}.}
    \label{table:main results_full}
    \resizebox{\textwidth}{!}{%

    \begin{tabular}{l r r r r r r r r} 
    \toprule
       & \multicolumn{2}{c}{\textbf{V*Bench}} & & \multicolumn{2}{c}{\textbf{HRBench-4K}} & & \multicolumn{2}{c}{\textbf{HRBench-8K}}\\
      \textbf{Model} & \textbf{\makecell{Overall Acc. $\uparrow$ \\(Attr | Spatial) [\%]} } & \textbf{\makecell{FP [\#]} $\downarrow$} & & \textbf{\makecell{Overall Acc. $\uparrow$ \\(FSP | FCP) [\%]} } & \textbf{\makecell{FP [\#]} $\downarrow$} & & \textbf{\makecell{Overall Acc. $\uparrow$\\(FSP | FCP) [\%]}} & \textbf{\makecell{FP [\#]} $\downarrow$} \\

    \specialrule{.1em}{.1em}{.1em} 
        \rule{0pt}{\rowheight} {\large\textbf{ZoomEye}} & & & \vline & & & \vline & & \\
        
        \specialrule{.05em}{.1em}{.1em} 
        \textit{LLaVA-1.5-7B} & & & \vline & & & \vline & & \\ 
        \texttt{Depth-1 (2 crops)} & \makecell{50.26 \\ \color{gray}(41.74 | 63.16)\color{black}} & 12.50 & \vline & \makecell{36.25 \\ \color{gray}(39.25 | 33.25)\color{black}} & 11.54 & \vline & \makecell{33.88 \\ \color{gray}(32.25 | 35.5)\color{black}} & 11.55 \\
        \texttt{Depth-1 (4 crops)} & \makecell{50.78 \\ \color{gray}(41.74 | 64.47)\color{black}} & 20.37 & \vline & \makecell{39.13 \\ \color{gray}(44.75 | 33.5)\color{black}} & 17.46 & \vline & \makecell{32.88 \\ \color{gray}(32.75 | 33.00)\color{black}} & 18.18 \\
		\texttt{Depth-2 (4 crops)} & \makecell{71.20 \\ \color{gray}(67.83 | 76.32)\color{black}} & 44.54 & \vline & \makecell{47.75 \\ \color{gray}(57.25 | 38.25)\color{black}} & 35.60 & \vline & \makecell{42.13 \\ \color{gray}(49.00 | 36.25)\color{black}} & 39.15 \\
        \texttt{Depth-5 (4 crops)} & \makecell{\textbf{77.48} \\ \color{gray}(80.87 | 72.37)\color{black}} & 48.63 & \vline & \makecell{\textbf{50.00} \\ \color{gray}(63.25 | 36.75)\color{black}} & 49.38 & \vline & \makecell{\textbf{49.00} \\ \color{gray}(61.75 | 36.25)\color{black}} & 59.64 \\

        \specialrule{.05em}{.1em}{.1em} 
        \textit{LLaVA-OV-7B} & & & \vline & & & \vline & & \\ 
        \texttt{Depth-1 (2 crops)} & \makecell{75.92 \\ \color{gray}(77.39 | 73.68)\color{black}} & 10.52 & \vline & \makecell{61.38 \\ \color{gray}(70.00 | 52.75)\color{black}} & 9.46 & \vline & \makecell{59.00 \\ \color{gray}(66.50 | 51.50)\color{black}} & 9.02 \\
        \texttt{Depth-1 (4 crops)} & \makecell{81.15 \\ \color{gray}(81.74 | 80.26)\color{black}} & 13.34 & \vline & \makecell{65.75 \\ \color{gray}(80.75 | 50.75)\color{black}} & 11.43 & \vline & \makecell{59.63 \\ \color{gray}(69.00 | 50.25)\color{black}} & 12.59 \\
		\texttt{Depth-2 (4 crops)} & \makecell{90.05 \\ \color{gray}(93.04 | 85.53)\color{black}} & 21.08 & \vline & \makecell{66.13 \\ \color{gray}(80.25 | 52.00)\color{black}} & 20.42 & \vline & \makecell{65.63 \\ \color{gray}(81.75 | 49.5)\color{black}} & 22.31 \\
        \texttt{Depth-5 (4 crops)} & \makecell{\textbf{91.10} \\ \color{gray}(93.91 | 86.84)\color{black}} & 23.98 & \vline & \makecell{\textbf{69.38} \\ \color{gray}(84.5 | 54.25)\color{black}} & 29.19 & \vline & \makecell{\textbf{69.00} \\ \color{gray}(86.75 | 51.25)\color{black}} & 36.75 \\
    
    \specialrule{.1em}{.1em}{.1em} 
        \rule{0pt}{\rowheight} {\large\textbf{ViCrop}} & & & \vline & & & \vline & & \\ 
        
        \specialrule{.05em}{.1em}{.1em} 
        \textit{LLaVA-1.5-7B} & & & \vline & & & \vline & & \\ 
        \texttt{rel-att-high} & \makecell{\textbf{59.16} \\ \color{gray}(58.26 | 60.53)\color{black}} & 12.26 & \vline & \makecell{42.50 \\ \color{gray}(51.50 | 33.50)\color{black}} & 30.59 & \vline & \makecell{\textbf{39.38} \\ \color{gray}(48.00 | 30.75)\color{black}} & 78.60 \\
        \texttt{grad-att-high} & \makecell{54.97 \\ \color{gray}(53.04 | 57.90)\color{black}} & 6.63 & \vline & \makecell{\textbf{44.25} \\ \color{gray}(53.75 | 34.75)\color{black}} & 15.80 & \vline & \makecell{38.38 \\ \color{gray}(44.25 | 32.50)\color{black}} & 39.80 \\
    
    \specialrule{.1em}{.1em}{.1em} 
        {\large\textbf{DC\textsuperscript{2}}} & & & \vline & & & \vline & & \\ 
        
        \specialrule{.05em}{.1em}{.1em} 
        \textit{LLaVA-v.1.5-7B} & \makecell{\textbf{57.00} \\ \color{gray}(----- | -----)\color{black}} & 18.18 & \vline & \makecell{\textbf{42.30} \\ \color{gray}(----- | -----)\color{black}} & 48.55 & \vline & \makecell{\textbf{39.50} \\ \color{gray}(----- | -----)\color{black}} & 77.02 \\

    \specialrule{.1em}{.15em}{.15em} 
        \rule{0pt}{\rowheight} {\large\textbf{SEAL}} & \makecell{\textbf{73.68} \\ \color{gray}(----- | -----)\color{black}} & 25.53 & \vline & \makecell{\textbf{34.50} \\ \color{gray}(----- | -----)\color{black}} & 18.05 & \vline & \makecell{\textbf{33.50} \\ \color{gray}(----- | -----)\color{black}} & 16.96 \\

    \specialrule{.1em}{.1em}{.1em} 
        \rule{0pt}{\rowheight} {\large\textbf{\texttt{FOCUS} (Ours)}}& & & \vline & & & \vline & & \\ 
        
        \specialrule{.05em}{.1em}{.1em} 
\textit{LLaVA-1.5-7B} & & & \vline & & & \vline & & \\ 

\texttt{Steps-1 (no-overrun)} & \makecell{51.30 \\ \color{gray}(46.95 | 57.89)\color{black}} & 1.47 & \vline & \makecell{41.13 \\ \color{gray}(49.75 | 32.5)\color{black}} & 3.14 & \vline & \makecell{40.63 \\ \color{gray}(46.00 | 35.25)\color{black}} & 3.14 \\
\texttt{Steps-2 (no-overrun)} & \makecell{57.07 \\ \color{gray}(53.91 | 61.84)\color{black}} & 4.25 & \vline & \makecell{46.5 \\ \color{gray}(56.00 | 37.00)\color{black}} & 5.41 & \vline & \makecell{40.63 \\ \color{gray}(44.75 | 36.50)\color{black}} & 5.41 \\
\texttt{Steps-1 (overrun)} & \makecell{64.40 \\ \color{gray}(63.48 | 65.79)\color{black}} & 4.86 & \vline & \makecell{42.25 \\ \color{gray}(51.50 | 33.00)\color{black}} & 4.99 & \vline & \makecell{42.38 \\ \color{gray}(48.50 | 36.25)\color{black}} & 5.08 \\
\texttt{Steps-2 (overrun)} & \makecell{66.49 \\ \color{gray}(66.09 | 67.11)\color{black}} & 5.70 & \vline & \makecell{45.88 \\ \color{gray}(55.75 | 36.00)\color{black}} & 5.95 & \vline & \makecell{42.15 \\ \color{gray}(46.75 | 37.50)\color{black}} & 6.04 \\
\texttt{Steps-3 (overrun)} & \makecell{67.01 \\ \color{gray}(66.09 | 68.42)\color{black}} & 6.79 & \vline & \makecell{47.13 \\ \color{gray}(56.50 | 37.75)\color{black}} & 9.09 & \vline & \makecell{44.13 \\ \color{gray}(48.00 | 40.25)\color{black}} & 9.07 \\
\texttt{Steps-4 (overrun)} & \makecell{68.06 \\ \color{gray}(66.96 | 69.74)\color{black}} & 8.27 & \vline & \makecell{49.25 \\ \color{gray}(59.75 | 38.75)\color{black}} & 10.14 & \vline & \makecell{\textbf{45.00} \\ \color{gray}(50.25 | 39.75)\color{black}} & 10.10 \\
\texttt{Steps-6 (overrun)} & \makecell{70.68 \\ \color{gray}(70.43 | 71.05)\color{black}} & 10.71 & \vline & \makecell{50.63 \\ \color{gray}(62.25 | 39.00)\color{black}} & 12.31 & \vline & \makecell{\textbf{45.00} \\ \color{gray}(52.00 | 38.00)\color{black}} & 12.23 \\
\texttt{Steps-8 (overrun)} & \makecell{\textbf{72.77} \\ \color{gray}(72.17 | 73.68)\color{black}} & 13.28 & \vline & \makecell{\textbf{51.75} \\ \color{gray}(64.00 | 39.50)\color{black}} & 14.49 & \vline & \makecell{44.13 \\ \color{gray}(52.25 | 36.00)\color{black}} & 14.41 \\

        \specialrule{.05em}{.1em}{.1em} 
        \textit{LLaVA-OV-7B} & & & \vline & & & \vline & & \\ 
        \texttt{Steps-1 (no-overrun)} & \makecell{83.24 \\ \color{gray}(87.82 | 76.31)\color{black}} & 1.47 & \vline & \makecell{69.00 \\ \color{gray}(82.25 | 55.75)\color{black}} & 3.84 & \vline & \makecell{65.75 \\ \color{gray}(77.00 | 54.50)\color{black}} & 3.86 \\
\texttt{Steps-2 (no-overrun)} & \makecell{89.01 \\ \color{gray}(92.17 | 84.21)\color{black}} & 4.23 & \vline & \makecell{70.38 \\ \color{gray}(84.00 | 56.75)\color{black}} & 6.07 & \vline & \makecell{66.88 \\ \color{gray}(78.00 | 55.75)\color{black}} & 6.09 \\
\texttt{Steps-1 (overrun)} & \makecell{90.57 \\ \color{gray}(93.04 | 86.84)\color{black}} & 4.05 & \vline & \makecell{69.88 \\ \color{gray}(85.75 | 54.00)\color{black}} & 6.55 & \vline & \makecell{68.75 \\ \color{gray}(81.00 | 56.50)\color{black}} & 7.35 \\
\texttt{Steps-2 (overrun)} & \makecell{91.62 \\ \color{gray}(93.93 | 88.15)\color{black}} & 5.16 & \vline & \makecell{70.25 \\ \color{gray}(86.50 | 54.00)\color{black}} & 7.41 & \vline & \makecell{67.63 \\ \color{gray}(79.00 | 56.25)\color{black}} & 8.06 \\
\texttt{Steps-3 (overrun)} & \makecell{91.62 \\ \color{gray}(93.91 | 88.15)\color{black}} & 6.37 & \vline & \makecell{70.00 \\ \color{gray}(85.75 | 54.25)\color{black}} & 8.34 & \vline & \makecell{66.88 \\ \color{gray}(78.00 | 55.75)\color{black}} & 8.93 \\
\texttt{Steps-4 (overrun)} & \makecell{\textbf{92.15} \\ \color{gray}(93.91 | 89.47)\color{black}} & 7.63 & \vline & \makecell{70.75 \\ \color{gray}(85.75 | 55.75)\color{black}} & 9.32 & \vline & \makecell{68.38 \\ \color{gray}(80.75 | 56.00)\color{black}} & 9.86 \\
\texttt{Steps-6 (overrun)} & \makecell{\textbf{92.15} \\ \color{gray}(93.91 | 89.47)\color{black}} & 10.22 & \vline & \makecell{70.63 \\ \color{gray}(85.75 | 55.50)\color{black}} & 11.33 & \vline & \makecell{68.88 \\ \color{gray}(82.50 | 55.25)\color{black}} & 11.81 \\
\texttt{Steps-8 (overrun)} & \makecell{91.62 \\ \color{gray}(93.04 | 89.47)\color{black}} & 12.84 & \vline & \makecell{\textbf{71.13} \\ \color{gray}(86.75 | 55.50)\color{black}} & 13.41 & \vline & \makecell{\textbf{69.63} \\ \color{gray}(83.50 | 55.75)\color{black}} & 13.83 \\
        
    \bottomrule
    \end{tabular}%
  }
  \end{center}
\end{table*}
\begin{table*}[hbtp]
\footnotesize
    \begin{center}
    \caption{\textbf{Detailed results on the MME-RealWorld-Lite dataset.} Accuracy is reported both per subtask and on average per subset. The highest accuracy for each subtask is highlighted in bold.}
    \label{table:mme-real-world-full}
    \begin{tabular}{l l l c c c c} 
        \toprule
        \multicolumn{3}{c}{\textbf{Task}} & & \multicolumn{3}{c}{\textit{LLaVA-OV-7B}}\\
        & & & & \makecell{\textbf{Vanilla} \\ \textbf{Acc.} $[\%]$} & \makecell{\textbf{ZoomEye} \\ \textbf{Acc.} $[\%]$} & \makecell{\texttt{FOCUS} \textbf{(Ours)} \\ \textbf{Acc.} $[\%]$}\\
        \midrule
        \multirow{26}{*}{\rotatebox{90}{Perception}} & \multirow{7}{*}{AD} & \texttt{Motion\textsubscript{multi-pedestrians}} & \vline & 22.00 & 28.00 & \textbf{34.00}\\
        & & \texttt{Motion\textsubscript{multi-vehicles}} & \vline & \textbf{46.00} & 38.00 & 40.00\\ 
        & & \texttt{Motion\textsubscript{pedestrian}} & \vline & 24.00 & \textbf{46.00} & 44.00\\
        & & \texttt{Motion\textsubscript{vehicle}} & \vline & 24.00 & \textbf{54.00} & 34.00\\
        & & \texttt{Object\textsubscript{count}} & \vline & 36.00 & \textbf{42.00} & 40.00\\
        & & \texttt{Object\textsubscript{identify}} & \vline & \textbf{78.00} & 74.00 & 70.00\\
        & & \texttt{Visual\textsubscript{traffic-signal}} & \vline & 62.00 & \textbf{78.00} & 68.00\\ \cline{2-7}
        & \multirow{2}{*}{DT} & \texttt{Diagram} & \vline & 70.00 & \textbf{80.00} & 60.00\\
        & & \texttt{Table} & \vline & 60.00 & \textbf{68.00} & 56.00\\ \cline{2-7}
        & \multirow{7}{*}{MO} & \texttt{Person\textsubscript{color}} & \vline & 32.00 & 40.00 & \textbf{56.00}\\
        & & \texttt{Person\textsubscript{counting}} & \vline & 32.00 & \textbf{38.00} & \textbf{38.00}\\
        & & \texttt{Person\textsubscript{orientation}} & \vline & 10.53 & \textbf{15.79} & 10.53\\
        & & \texttt{Vehicle\textsubscript{color}} & \vline & 46.00 & \textbf{60.00} & 52.00\\
        & & \texttt{Vehicle\textsubscript{counting}} & \vline & 56.00 & 56.00 & \textbf{58.00}\\
        & & \texttt{Vehicle\textsubscript{location}} & \vline & \textbf{38.00} & 28.00 & 28.00\\
        & & \texttt{Vehicle\textsubscript{orientation}} & \vline & 12.00 & 20.00 & \textbf{26.00}\\ \cline{2-7}
        & \multirow{5}{*}{OCR} & \texttt{Advert \& product} & \vline & 82.00 & 82.00 & \textbf{88.00}\\
        & & \texttt{Book map poster} & \vline & \textbf{78.00} & 70.00 & 74.00\\
        & & \texttt{License} & \vline & \textbf{88.00} & 86.00 & \textbf{88.00}\\
        & & \texttt{Phone \& address} & \vline & 82.00 & \textbf{96.00} & 90.00\\
        & & \texttt{Text recognition} & \vline & 78.00 & 72.00 & \textbf{80.00}\\ \cline{2-7}
        & \multirow{3}{*}{RS} & \texttt{Color} & \vline & 60.00 & \textbf{64.00} & \textbf{64.00}\\
        & & \texttt{Count} & \vline & 34.00 & \textbf{40.00} & 20.00\\
        & & \texttt{Position} & \vline & \textbf{62.00} & 50.00 & 56.00\\ \cmidrule{2-7}\morecmidrules\cmidrule{2-7}
        & \multicolumn{2}{r}{\textbf{Average}} & \vline & 52.01& \textbf{56.29} & 54.15\\
        \midrule
        \multirow{17}{*}{\rotatebox{90}{Reasoning}} & \multirow{8}{*}{AD} & \texttt{Attention\textsubscript{traffic-signal}} & \vline & \textbf{74.00} & 72.00 & \textbf{74.00}\\
        & & \texttt{Intention\textsubscript{ego}} & \vline & \textbf{26.00} & 24.00 & 22.00\\
        & & \texttt{Intention\textsubscript{pedestrian}} & \vline & 48.00 & 50.00 & \textbf{54.00}\\
        & & \texttt{Intention\textsubscript{vehicle}} & \vline & 26.00 & \textbf{42.00} & 40.00\\
        & & \texttt{Interaction\textsubscript{ego-2-pedestrian}} & \vline & 20.00 & \textbf{28.00} & 22.00\\
        & & \texttt{Interaction\textsubscript{ego-2-traffic-signal}} & \vline & 28.00 & \textbf{30.00} &22.00\\
        & & \texttt{Interaction\textsubscript{ego-2-vehicle}} & \vline & 26.00 & 22.00 & \textbf{26.00}\\
        & & \texttt{Interaction\textsubscript{other-2-other}} & \vline & 10.00 & \textbf{12.00} & 8.00\\ \cline{2-7}
        & \multirow{2}{*}{DT} & \texttt{Diagram} & \vline & 40.00 & 54.00 & \textbf{58.00}\\
        & & \texttt{Table} & \vline & 40.00 & 44.00 & \textbf{46.00}\\ \cline{2-7}
        & \multirow{3}{*}{MO} & \texttt{Calculate} & \vline & 42.00 & 46.00 & \textbf{50.00}\\
        & & \texttt{Intention} & \vline & 26.00 & 26.00 & \textbf{42.00}\\
        & & \texttt{Property} & \vline & 64.00 & \textbf{66.00} & 62.00\\ \cline{2-7}
        & \multirow{2}{*}{OCR} & \texttt{Character identification} & \vline & 72.00 & 64.00 & \textbf{74.00}\\
        & & \texttt{Scene understanding} & \vline & \textbf{72.00} & 64.00 & 68.00\\ \cmidrule{2-7}\morecmidrules\cmidrule{2-7}
        & \multicolumn{2}{r}{\textbf{Average}} & \vline & 40.93& 43.20 & \textbf{44.53}\\
        \bottomrule
    \end{tabular}
  \end{center}
\end{table*}

\subsection{Open-ended VQA}
\label{appendix:open_VQA}

\rebuttal{
While the primary results in the main paper are concerned with multiple-choice \ac{VQA}, here we focus on the evaluation on open-ended \ac{VQA}.
As we are not aware of fine-grained datasets focusing on open-ended \ac{VQA}, we reuse V*Bench for this task.
While it follows the multiple-choice format, it also provides the ground-truth answer in a natural language format, e.g., \texttt{"The color of the dog is white."}.
To explore the open-ended \ac{VQA} capabilities of \texttt{FOCUS} with LLaVA-1.5, we provide it with \ac{VQA} questions without answer options, e.g., \texttt{"What is the color of the dog?"}.
Then, we compare the responses with the ground-truth sequence
using an \ac{LLM}-as-a-judge framework \cite{llmasajudge_zheng_2023}, leveraging \textit{Qwen-2.5-7B} \cite{qwen_2_5_vl_bai}.
Moreover, we manually review Qwen-2.5’s judgments and correct any misclassifications.
The results of this analysis for LLaVA-1.5 clearly show that \texttt{FOCUS} substantially improves the fine-grained open-ended VQA performance, increasing accuracy from $44.50\%$ for the vanilla LLaVA-1.5 model to $65.97\%$.}

\subsection{Analysis of hyperparameter influence}
\label{subsect:full_robustness_analysis}

\rebuttal{We further investigate how variations in the hyperparameters of \texttt{FOCUS} affect its performance.
This analysis is conducted using \texttt{FOCUS} with LLaVA-1.5 and LLaVA-OV on V*Bench, focusing on five key hyperparameters, namely the number of anchor points ($k$), the ROI expansion threshold, the maximum ROI size ($s_{\text{max}}$), the minimal Euclidean distance between anchor points ($s_{\text{dist}}$) and the NMS threshold.
As shown in \Cref{tab:hyperparameter_analysis_main}, 
using alternative hyperparameter settings reduces accuracy by at most $4.7$ pp. compared to the original configuration,
demonstrating our method's low sensitivity to hyperparameters.
During the sensitivity analysis, we discover some alternative hyperparameter configurations that achieve even higher accuracy than our default settings.}

\begin{table}[h]
\centering
\footnotesize
\caption{\textbf{Hyperparameter analysis of \texttt{FOCUS}.} We assess the impact of five hyperparameters on the performance of \texttt{FOCUS}. ${}^\dagger$ indicates the original hyperparameter.}
\label{tab:hyperparameter_analysis_main}

\begin{minipage}[t]{0.48\textwidth}
    \begin{subtable}[t]{\linewidth}
        \caption{Variation of numbers of anchor points $k$.}
        \label{tab:13a}
        \begin{tabularx}{\linewidth}{L{0.2} C{0.33} C{0.33}}
            \toprule
            \textbf{Model} & \textbf{$k$} & \textbf{Accuracy $[\%]$} \\
            \midrule
            \multirow{2}{*}{LLaVA-1.5} & $30^\dagger$ & $72.77$ \\
             & $\sim\mathcal{U}(10,50)$ & $72.77 \pm 1.55$\\
            \midrule
            \multirow{2}{*}{LLaVA-OV} & $30^\dagger$ & $92.15$ \\
             & $\sim\mathcal{U}(10,50)$ & $92.03 \pm 1.37$\\
            \bottomrule
        \end{tabularx}
    \end{subtable}

    \vspace{1em}

    \begin{subtable}[t]{\linewidth}
        \caption{Variation of ROI expansion threshold.}
        \label{tab:13b}
        \begin{tabularx}{\linewidth}{L{0.2} C{0.33} C{0.33}}
            \toprule
            \textbf{Model} & \textbf{\makecell{ROI expans. \\ threshold}} & \textbf{Accuracy $[\%]$} \\
            \midrule
            \multirow{2}{*}{LLaVA-1.5} & $0.5^\dagger$ & $72.77$ \\
             & $\sim\mathcal{U}(0.3,0.7)$ & $72.77 \pm 0.00$\\
            \midrule
            \multirow{2}{*}{LLaVA-OV} & $0.5^\dagger$ & $92.15$ \\
             & $\sim\mathcal{U}(0.3,0.7)$ & $92.70 \pm 0.25$\\
            \bottomrule
        \end{tabularx}
    \end{subtable}

    \vspace{1em}

    \begin{subtable}[t]{\linewidth}
        \caption{Variation of maximum ROI size $s_{\text{max}}$.}
        \label{tab:13c}
        \begin{tabularx}{\linewidth}{L{0.2} C{0.33} C{0.33}}
            \toprule
            \textbf{Model} & \textbf{$s_{\text{max}}$} & \textbf{Accuracy $[\%]$} \\
            \midrule
            \multirow{3}{*}{LLaVA-1.5} & $\;\,5^\dagger$ & $72.77$ \\
             & $7$ & $69.63$\\
             & $9$ & $68.06$\\
            \midrule
            \multirow{4}{*}{LLaVA-OV} & $\;\, 9^\dagger$ & $92.15$ \\
             & $5$ & $94.24$\\
             & $7$ & $91.01$\\
             & $11$ & $89.53$\\
            \bottomrule
        \end{tabularx}
    \end{subtable}
\end{minipage}
\hfill
\begin{minipage}[t]{0.48\textwidth}
    \begin{subtable}[t]{\linewidth}
        \caption{Variation of minimum distance between ROI anchor points $s_{\text{dist}}$.}
        \label{tab:13d}
        \begin{tabularx}{\linewidth}{L{0.2} C{0.33} C{0.33}}
            \toprule
            \textbf{Model} & \textbf{${s}_{\text{dist}}$} & \textbf{Accuracy $[\%]$} \\
            \midrule
            \multirow{3}{*}{LLaVA-1.5} & $\;\, 2^\dagger$ & $72.77$ \\
             & $3$ & $72.25$\\
             & $4$ & $69.11$\\
            \midrule
            \multirow{4}{*}{LLaVA-OV} & $\;\, 2^\dagger$ & $92.15$ \\
             & $3$ & $91.62$\\
             & $4$ & $92.15$\\
             & $5$ & $91.62$\\
            \bottomrule
        \end{tabularx}
    \end{subtable}

    \vspace{1em}

    \begin{subtable}[t]{\linewidth}
        \caption{Variation of NMS threshold.}
        \label{tab:13e}
        \begin{tabularx}{\linewidth}{L{0.2} C{0.33} C{0.33}}
            \toprule
            \textbf{Model} & \textbf{\makecell{NMS \\ threshold}} & \textbf{Accuracy $[\%]$} \\
            \midrule
            \multirow{4}{*}{LLaVA-1.5} & $\;\, 0.3^\dagger$ & $72.77$ \\
             & $0.1$ & $70.16$\\
             & $0.5$ & $72.25$\\
             & $0.7$ & $72.25$\\
            \midrule
            \multirow{4}{*}{LLaVA-OV} & $\;\, 0.3^\dagger$ & $92.15$ \\
             & $0.1$ & $92.15$\\
             & $0.5$ & $92.15$\\
             & $0.7$ & $92.15$\\
            \bottomrule
        \end{tabularx}
    \end{subtable}
\end{minipage}

\end{table}

\rebuttal{
For $k$ and the \ac{ROI} expansion threshold, we randomly sample 50 values from $\mathcal{U}(10, 50)$ and $\mathcal{U}(0.3, 0.7)$, respectively, where $\mathcal{U}$ indicates a uniform distribution.
Across both LLaVA-1.5 and LLaVA-OV, we observe only minor performance variations when adjusting $k$: an accuracy of $72.77 \pm 1.55$ for LLaVA-1.5 and $92.03 \pm 1.37$ for LLaVA-OV, as shown in \Cref{tab:13a}.
For the \ac{ROI} expansion threshold, we observe an even smaller impact on the performance of FOCUS, with an accuracy of $72.77 \pm 0.00$ for LLaVA-1.5 and $92.70 \pm 0.25$ for LLaVA-OV, as shown in \Cref{tab:13b}.
For $s_{\text{dist}}$, $s_{\text{max}}$, and the NMS threshold, we vary the values in discrete steps and analyze their influence on performance.
Across both LLaVA-1.5 and LLaVA-OV, we find that $s_{\text{max}}$ has the largest impact on \texttt{FOCUS} among all analyzed hyperparameters, resulting in an accuracy drop of $4.71$ pp. for LLaVA-1.5 and $2.62$ pp. for LLaVA-OV, as shown in \Cref{tab:13c}.
Interestingly, for \texttt{FOCUS} with LLaVA-OV, we observe an accuracy improvement of $2.09$ pp. over the baseline when setting $s_{\text{max}} = 5$.
For $s_{\text{dist}}$, we observe a maximum accuracy degradation of $3.66$ pp. for LLaVA-1.5 and $0.53$ pp. for LLaVA-OV, as shown in \Cref{tab:13d}.  
For the NMS threshold, we find only a minor impact, with a maximum accuracy degradation of $2.61$ pp. for LLaVA-1.5 and no observable influence for LLaVA-OV, as shown in \Cref{tab:13e}. This is likely because FOCUS with LLaVA-OV generates higher-resolution object relevance maps, reducing its reliance on NMS.
}

\subsection{Additional efficiency metrics}
\label{appendix:efficiency_measurements}
We report additional efficiency metrics—including average execution time and peak GPU memory usage per sample—on V*Bench.
We compare SEAL with a customized LLaVA-7B, vanilla LLaVA-1.5-7B, vanilla LLaVA-OneVision-7B, and three training-free methods (ViCrop, ZoomEye, and \texttt{FOCUS}) using \mbox{LLaVA-1.5}~models.

As shown in \Cref{tab:add_efficiency}, among the training-free methods, ZoomEye achieves the highest accuracy but suffers from poor efficiency due to its complex confidence prediction mechanism, which involves multiple question prompts and a hierarchical tree structure.
This is reflected by its high number of \acfp{FP} and long execution times. \texttt{FOCUS}, by contrast, leverages an object relevance map built from cached token similarities to directly identify relevant image regions.
As a result, it requires only 25\% of ZoomEye’s \acp{FP} and execution time to reach comparable accuracy.
ViCrop is slightly more efficient than \texttt{FOCUS} in terms of execution time and \acp{FP}, but it achieves lower accuracy and incurs the highest peak memory usage due to its incompatibility with efficient attention mechanisms.
SEAL differs architecturally from LLaVA-1.5 and LLaVA-OneVision.
Its dual-MLLM design makes it significantly slower and more memory-intensive than most methods in the comparison.
In general, a lower number of \acp{FP} is associated with reduced execution time.

\begin{table*}[hbtp]
\caption{\textbf{Additional performance and efficiency metrics on V*Bench.}  In the last three rows, the best-performing method is highlighted in bold and the runner-up is underlined.}
\label{tab:add_efficiency}
\centering
\renewcommand{\arraystretch}{1.2}
\resizebox{0.95\textwidth}{!}{
\begin{tabular}{l c c c c} 
    \toprule
    \textbf{Model} &  \textbf{Acc. $[\%]$$\uparrow$}  &  \textbf{FP  $[\#]$$\downarrow$}  & \textbf{Avg. execution time $[s]$$\downarrow$}  & \textbf{Avg. peak memory $[GB]$$\downarrow$} \\
    \midrule
    \textbf{SEAL} & 73.68 & 25.53 & 9.16 & 27.34 \\
    \textbf{LLaVA-OV-7B} & 74.46 & - & 1.30 & 19.64 \\
    \hline
    \textbf{LLaVA-1.5-7B} & 47.64 & - & 0.25 & 13.57 \\
    + w/ ViCrop & 59.16 & \textbf{12.26} & \textbf{1.36}& 19.93 \\
    + w/ ZoomEye & \textbf{77.49} & 48.63 & 11.26 & \textbf{14.24} \\
    + w/ Ours & \underline{72.77} & \underline{13.28} & \underline{2.19} & \underline{14.91} \\
    \bottomrule
\end{tabular}
}
\end{table*}

\rebuttal{
Additionally, we compare the efficiency of \texttt{FOCUS} and ZoomEye with LLaVA-1.5 across multiple configurations (see \Cref{tab:combined_config}).
We evaluate them on V*Bench in terms of accuracy, average inference time, average \acp{FP} and average FLOPs.
The latter ones are estimated based on a calculation scheme applied in prior \ac{MLLM} work \cite{chen_fastv_2024, xing_pyramid_2025}.
We ran the evaluations on the same hardware to ensure result~comparability.
}

\rebuttal{
Notably, the lowest-complexity configuration of ZoomEye exhibits a higher inference time and nearly identical \acp{FP} and FLOPs compared to the highest-complexity configuration of \texttt{FOCUS}.  
Despite this, \texttt{FOCUS} outperforms ZoomEye by $22.51$ pp. in accuracy under this configuration.
Moreover, ZoomEye shows a steep increase in complexity—measured by inference time, \acp{FP}, and FLOPs—as the search depth in its tree structure increases.  
Its maximum configuration achieves an accuracy of $77.48\%$, with an inference time of $11.96$ seconds, $48.63$ \acp{FP}, and a computational cost of $217$ TFLOPs.
}

\rebuttal{
In summary, the efficiency gains reported in terms of \acp{FP} are consistently reflected in inference time and FLOPs.
\texttt{FOCUS} achieves competitive or superior accuracy with significantly lower inference time and fewer FLOPs compared to ZoomEye.
For instance, at \texttt{Steps-6 (overrun)}, \texttt{FOCUS} reaches $70.68\%$ accuracy in just $2.00$ seconds and $51.26$ TFLOPs, whereas ZoomEye (\texttt{Depth-2}) requires nearly $4\times$ more FLOPs and $5\times$ longer inference time to reach a comparable accuracy.
These results underscore the efficiency of \texttt{FOCUS} for fine-grained visual reasoning tasks in practice.
}

\begin{table*}[h!]
    \centering
    \caption{\textbf{In-depth efficiency comparison of \texttt{FOCUS} and ZoomEye.} Across different configurations, we compare efficiency in terms of accuracy, average inference time, average \acp{FP} and average FLOPs on V*Bench.}
    \label{tab:combined_config}
    \resizebox{0.95\textwidth}{!}{%
    \begin{tabular}{l c c c c}
        \toprule
        \textbf{Model} & \textbf{Accuracy $[\%] \uparrow$} & \textbf{Inference time $[s] \downarrow$} & \textbf{FP $[\#] \downarrow$} & \textbf{TFLOPs $[\#] \downarrow$} \\
        \midrule
        \multicolumn{5}{l}{ZoomEye} \\
        \midrule
        \textit{LLaVA-1.5-7B} & & & & \\
        \texttt{Depth-1 (2 crops)} & 50.26 & 3.78 & 12.50 & 55.95 \\
        \texttt{Depth-1 (4 crops)} & 50.78 & 4.76 & 20.37 & 91.03 \\
        \texttt{Depth-2 (4 crops)} & 71.20 & 9.73 & 44.54 & 199.21 \\
        \texttt{Depth-5 (4 crops)} & 77.48 & 11.96 & 48.63 & 217.00 \\
        \midrule
        \multicolumn{5}{l}{\textbf{\texttt{FOCUS}}} \\
        \midrule
        \textit{LLaVA-1.5-7B} & & & & \\
        \texttt{Steps-1 (no overrun)} & 51.30 & 0.99 & 1.47 & 10.98 \\
        \texttt{Steps-2 (no overrun)} & 57.07 & 1.28 & 4.25 & 23.11 \\
        \texttt{Steps-1 (overrun)} & 64.40 & 1.36 & 4.86 & 25.73 \\
        \texttt{Steps-2 (overrun)} & 66.49 & 1.44 & 5.70 & 29.43 \\
        \texttt{Steps-3 (overrun)} & 67.01 & 1.55 & 6.79 & 34.15 \\
        \texttt{Steps-4 (overrun)} & 68.06 & 1.73 & 8.27 & 40.61 \\
        \texttt{Steps-6 (overrun)} & 70.68 & 2.00 & 10.71 & 51.26 \\
        \texttt{Steps-8 (overrun)} & 72.77 & 2.27 & 13.28 & 62.46 \\
        \bottomrule
    \end{tabular}%
    }

\end{table*}

\subsection{Inference scheme and performance discrepancy}
\label{subsect:infer_scheme}
We describe in the following the comparison between different inference schemes and the discrepancy between reported and reproduced performance.

\begin{table*}[hbtp]
    \centering
    \caption{\textbf{Comparison of different inference schemes across models.} Each result includes accuracy, execution time, and number of forward passes for the visual search (FP). The highest accuracy in a column is highlighted in bold.}
    \label{tab:infer_scheme}
    \resizebox{0.95\textwidth}{!}{%
    \begin{tabular}{lccccccc}
        \toprule
        & \textbf{Vanilla} 
        & \multicolumn{3}{c}{\textbf{ZoomEye}} 
        & \multicolumn{3}{c}{\textbf{\texttt{FOCUS} (Ours)}} \\
        \cmidrule(lr){2-2} \cmidrule(lr){3-5} \cmidrule(lr){6-8}
        \textbf{Inference Scheme} 
        & \textbf{Acc. $[\%]$$\uparrow$} 
        & \textbf{Acc. $[\%]$$\uparrow$} 
        & \makecell{\textbf{Exec.} \\ \textbf{time}$[s]$}$\downarrow$
        & \textbf{FP $[\#] \downarrow$} 
        & \textbf{Acc. $[\%]$$\uparrow$} 
        & \makecell{\textbf{Exec.} \\ \textbf{time}$[s]$}$\downarrow$
        & \textbf{FP $[\#] \downarrow$} \\
        \midrule
        Logits matching         & \textbf{48.16} & \textbf{77.49} & 11.26 & 48.63 & \textbf{74.35} & 2.76 & 13.28 \\
        Open-ended generation  & 47.64          & 72.25          & 9.26  & 36.94 & 72.77          & 2.19 & 13.28 \\
        \bottomrule
    \end{tabular}}
\end{table*}

\textbf{Inference scheme: logits matching vs. open-ended generation}\hspace{1.8mm}
Multiple-choice \ac{VQA} requires selecting one of several fixed answer options. A common approach is open-ended generation \cite{vqa_antol, vqa_v2_goyal}, where the prompt includes the question and options (e.g., \texttt{"(A) Red"}), and the \ac{MLLM} generates the corresponding option letter.
In contrast, SEAL \cite{vstar_wu24} and ZoomEye \cite{zoomeye_shen2024} adopt an alternative scheme called logits matching on V*Bench \cite{vstar_wu24}. In this method, the model is prompted multiple times, once for each answer option. Specifically, each answer option is reformulated as a sentence (e.g., \texttt{"(A) Red"} → \texttt{"The color of the car is red."}) which is then appended to the original question. The model is prompted with these reformulated question–option pairs and the image, and the answer option yielding the highest logit score for its target tokens is selected as the final~prediction.

We noticed that ZoomEye utilizes logits matching on V*Bench but uses open-ended generation on the other three datasets, prompting us to investigate the impact of different inference schemes. We evaluate LLaVA-1.5 on V*Bench using both open-ended generation and logits matching across three methods: the vanilla baseline, ZoomEye, and \texttt{FOCUS} (ours). SEAL is excluded from this comparison, as it is not a training-free method. All other hyperparameters are kept constant.

As shown in \Cref{tab:infer_scheme}, both the vanilla model and \texttt{FOCUS} achieve $0.5$ pp. and $1.6$ pp. higher accuracy, respectively, when using logits matching.
This improvement likely stems from the fact that logits matching eliminates the need for strong instruction-following: models no longer need to explicitly generate the option letter, but instead compare the semantic content of full answer statements.
This makes the inference process more robust, particularly for models with weaker generative alignment.

In contrast, ZoomEye’s accuracy drops by over $5$ pp. and its execution time decreases by approximately $2$ seconds when switching to open-ended generation.
Given that \texttt{FOCUS} shows a $1.6$ pp. accuracy difference and a \mbox{$0.57$-second} reduction in execution time under the same scheme change, we attribute that portion of ZoomEye’s decline to the scheme itself.
The remaining gap—both in accuracy and runtime—can likely be attributed to suboptimal tuning in the open-ended setting, as all hyperparameters were held constant.
The changed inference mode likely alters the model confidence, which may lead to premature termination of tree search (indicated by the lower number of \acp{FP}) and increased prediction instability.

Despite the potential accuracy gains of logits matching, we adopt open-ended generation for \texttt{FOCUS} across all datasets.
This is done for two reasons.
First, the effectiveness of logits matching depends heavily on the quality of option reformulations, which are not always available or consistent across datasets.
This limits its generalizability and makes cross-dataset comparisons less reliable.
Second, logits matching requires one \acf{FP} per answer option, compared to only a single \ac{FP} in open-ended generation.
Although some acceleration of the logits matching scheme is implemented by caching question tokens, it still increases inference time (see \Cref{tab:infer_scheme}).
Therefore, we report \texttt{FOCUS} results using open-ended generation in the main paper.
For SEAL and ZoomEye, we preserve their original inference schemes.
Their respectively reported V*Bench performance is based on logits~matching.

\textbf{Discrepancy between reported and reproduced performance}\hspace{1.8mm}
For the baseline methods, i.e., SEAL \cite{vstar_wu24}, ViCrop \cite{vicrop_zhang2025}, and ZoomEye \cite{zoomeye_shen2024}, we use the official implementations provided in their respective repositories.
We strictly follow the original configurations, including software environments and data structures, as specified by each work.
As shown in \Cref{tab:reprod_acc}, we observe some discrepancies between the reported and reproduced performance, most notably for LLaVA-1.5 with ZoomEye.
We hypothesized this may be linked to the use of different efficient attention backends (e.g., FlashAttention-2 \cite{fa_dao} vs. PyTorch’s SDPA\footnote{See \url{https://docs.pytorch.org/tutorials/intermediate/scaled\_dot\_product\_attention\_tutorial.html}}).
However, the observed deviation with different attention implementations is small, less than $1\%$ on V*Bench—thus, further root causes of this deviation seem to exist, that however remain unclear.
To ensure fair and consistent comparisons under a unified evaluation setup, we always report the reproduced results for these benchmark methods in the main~paper.
\begin{table*}[t]
\centering
\caption{\textbf{Reported vs. reproduced accuracy across fine-grained \ac{VQA} datasets.} A dash indicates that no evaluation on that benchmark was performed in the original work.}
\label{tab:reprod_acc}
\resizebox{0.9\textwidth}{!}{
\begin{tabular}{lcccccc}
\toprule
& \multicolumn{2}{c}{\textbf{V*Bench}} 
& \multicolumn{2}{c}{\textbf{HRBench-4K}} 
& \multicolumn{2}{c}{\textbf{HRBench-8K}} \\
\cmidrule(lr){2-3} \cmidrule(lr){4-5} \cmidrule(lr){6-7}
\textbf{Model} 
& \makecell{\textbf{Reported} \\ \textbf{Acc.} $[\%]$} 
& \makecell{\textbf{Reprod.} \\ \textbf{Acc.} $[\%]$} 
& \makecell{\textbf{Reported} \\ \textbf{Acc.} $[\%]$} 
& \makecell{\textbf{Reprod.} \\ \textbf{Acc.} $[\%]$} 
& \makecell{\textbf{Reported} \\ \textbf{Acc.} $[\%]$} 
& \makecell{\textbf{Reprod.} \\ \textbf{Acc.} $[\%]$} \\
\midrule
\textbf{ZoomEye} &  &  &  &  &  & \\
\textit{LLaVA-1.5-7B}          & 83.25 & 77.48 & 53.25 & 50.00 & 50.75 & 49.00 \\
\textit{LLaVA-OV-7B}           & 90.58 & 91.10 & 69.63 & 69.38 & 69.25 & 69.00 \\
\midrule
\makecell[l]{\textbf{ViCrop}\\\textit{LLaVA-1.5-7B}} &  &  &  &  &  & \\
\texttt{rel-att-high}& 62.30 & 59.16 &   --  & 42.50 &   --  & 39.38 \\
\texttt{grad-att-high}& 57.07 & 54.97 &   --  & 44.25 &   --  & 38.38 \\
\midrule
\textbf{SEAL}                 & 75.39 & 73.68 &   --  & 34.50 &   --  & 33.50 \\
\bottomrule
\end{tabular}}
\end{table*}

\section{Further qualitative examples}
\label{sect:qualitative_examples}

\begin{figure}[hbtp]
    \centering
    \includegraphics[clip, trim=0cm 20.5cm 2.6cm 0cm, width=0.9\textwidth]{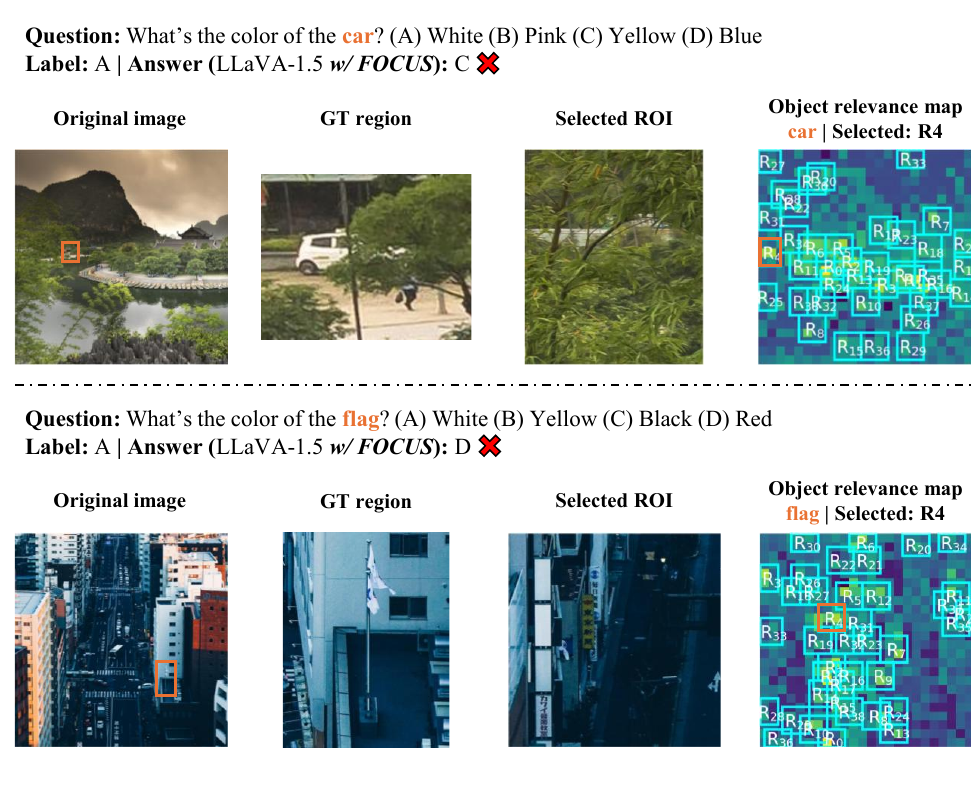}
    \caption{\textbf{Further failure cases of \texttt{FOCUS} with LLaVA-1.5.} We provide examples for failure cases of \texttt{FOCUS} using LLaVA-1.5 corresponding to the resolution limitation mentioned in \Cref{subsect:limitations}. Note that we manually highlight the relevant regions in the original image to facilitate easier localization of the ground truth area for the reader and these annotations are not included in the input for the \ac{MLLM}.}
    \label{fig:further_qualitative_examples_limitations}
\end{figure}

\begin{figure}[hbtp]
    \centering
    \includegraphics[clip, trim=0cm 7.0cm 1.5cm 0cm, width=0.9\textwidth]{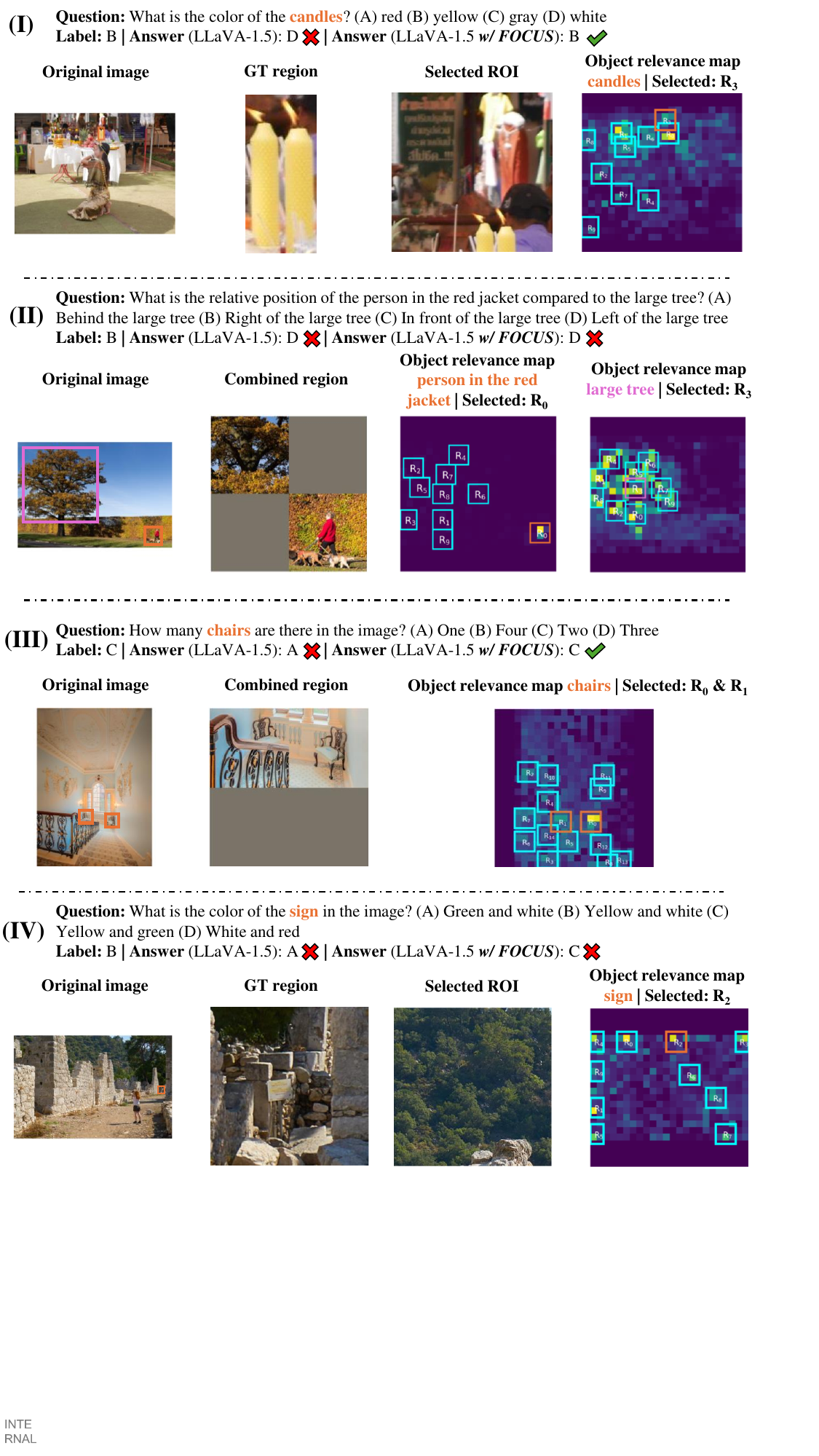}
    \caption{\textbf{Further qualitative examples of \texttt{FOCUS} with LLaVA-1.5.} We provide examples for single-object \textbf{(I)}, multi-object \textbf{(II)}, a type-2 question \textbf{(III)}, and a failure case \textbf{(IV)}. Note that we do not adjust the aspect ratio of the images for LLaVA-1.5. Therefore, there are some padding areas in the object relevance maps. Additionally, we manually highlight the relevant regions in the original image to facilitate easier localization of the ground truth area for the reader and these annotations are not included in the input for the \ac{MLLM}.}
    \label{fig:further_qualitative_examples_15}
\end{figure}

\begin{figure}[hbtp]
    \centering
    \includegraphics[clip, trim=0cm 7.0cm 1.5cm 0cm, width=0.9\textwidth]{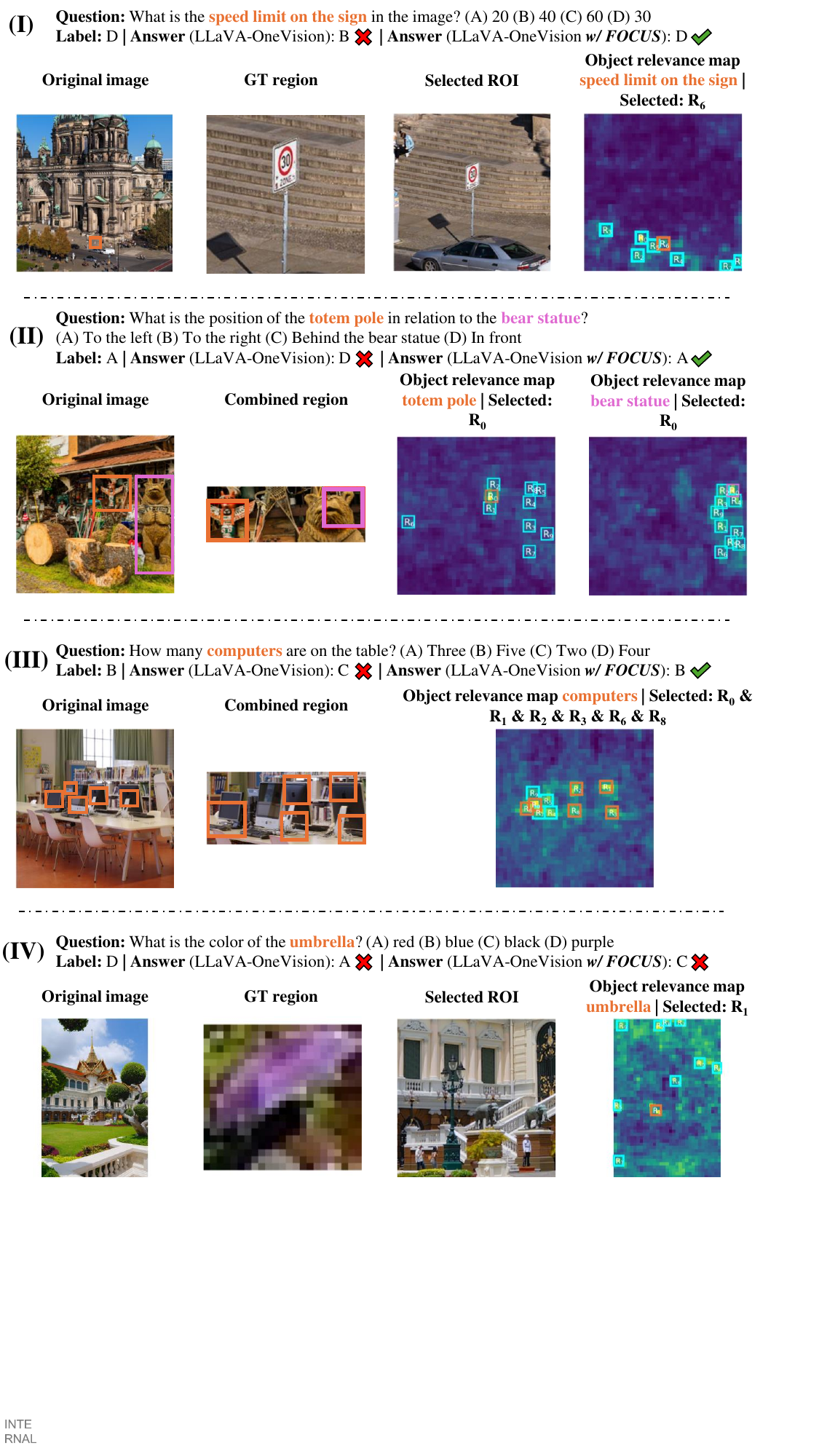}
    \caption{\textbf{Further qualitative examples of \texttt{FOCUS} with LLaVA-OneVision.} We provide examples for single-object \textbf{(I)}, multi-object \textbf{(II)}, a type-2 question \textbf{(III)}, and a failure case \textbf{(IV)}. Note that we manually highlight the relevant regions in the original image to facilitate easier localization of the ground truth area for the reader and these annotations are not included in the input for the \ac{MLLM}.}
    \label{fig:further_qualitative_examples_ov}
\end{figure}

This section provides additional qualitative examples that highlight both the strengths and limitations of \texttt{FOCUS} when applied to LLaVA-1.5 (see \Cref{fig:further_qualitative_examples_15}) and LLaVA-OneVision (see \Cref{fig:further_qualitative_examples_ov}).
For improved clarity in visualization, we use a reduced $k=10$, differing from the values specified in~\Cref{subsect:hyperpara}.
\rebuttal{Further, we provide qualitative examples for failure cases of \texttt{FOCUS} with LLaVA-1.5 on high-resolution images in \Cref{fig:further_qualitative_examples_limitations} to highlight the limitations of low-resolution object relevance maps (see \Cref{subsect:limitations}).
In these examples, we use an increased $k = 50$, enabling \texttt{FOCUS} to cover a larger portion of the image space; nevertheless, \texttt{FOCUS} still fails to identify the relevant image region.}

\Cref{fig:further_qualitative_examples_15} \textbf{(I)} showcases a type-1 single-target task with \texttt{FOCUS} and LLaVA-1.5.
By leveraging the \ac{MLLM}’s internal representations, \texttt{FOCUS} identifies a relevant crop that highlights the color of small candles, correcting the model’s initial \ac{VQA} response.
The \ac{ROI} ranking mechanism demonstrates robustness to noise in the object relevance map by assigning the highest confidence to the originally fourth-ranked region.
Moving to a type-1 multiple-target task, \Cref{fig:further_qualitative_examples_15} \textbf{(II)} illustrates how \texttt{FOCUS} identifies both the \texttt{person in the red jacket} and the \texttt{large tree} using in-context learning.
The object relevance maps clearly localize both targets.
Since LLaVA-1.5 accepts only single-image inputs, the selected \acp{ROI} are stitched together—a strategy detailed in \Cref{subsct:final_vqa_infer}—to avoid excessive image size or irrelevant content.
Despite this, the model fails to answer correctly, likely due to limited spatial reasoning.
A type-2 counting task is depicted in \Cref{fig:further_qualitative_examples_15} \textbf{(III)}.
Here, \texttt{FOCUS} successfully locates both chairs in the image.
As in the previous example, the regions are combined to form a single input for LLaVA-1.5.
This enables the model to correctly answer the \ac{VQA} query, which it could not do using the global view.
\Cref{fig:further_qualitative_examples_15} \textbf{(IV)}, finally, presents another failure case with LLaVA-1.5, underscoring the limitation discussed in \Cref{subsect:limitations}.
The example, drawn from the HRBench-8K dataset (see \Cref{tab:dataset_overview}), involves a high-resolution image where the sign is too small to be detected via the low-resolution object relevance map.
Consequently, \texttt{FOCUS} selects an incorrect region and fails to improve the \ac{VQA}~result.

Turning to LLaVA-OneVision, \Cref{fig:further_qualitative_examples_ov} \textbf{(I)} features a type-1 single-target task.
While vanilla LLaVA-OneVision fails to answer the question about the speed limit sign, \texttt{FOCUS} successfully identifies the relevant region using internal representations.
By isolating this region (see selected \ac{ROI}), the model is able to generate the correct \ac{VQA} response.
\Cref{fig:further_qualitative_examples_ov} \textbf{(II)} explores a type-1 multiple-target task.
Despite the relatively large size of the relevant regions, vanilla LLaVA-OneVision does not answer correctly.
\texttt{FOCUS} identifies the appropriate areas, generates a combined image region, and creates one local crop per relevant object.
The relevant regions are highlighted in the image of the combined \acp{ROI} with rectangles, helping reduce background noise and enabling the model to answer the \ac{VQA} task correctly.
Next, a type-2 counting task is shown in \Cref{fig:further_qualitative_examples_15} \textbf{(III)} with LLaVA-OneVision.
Vanilla LLaVA-OneVision fails to count the number of computers accurately.
\texttt{FOCUS} identifies five computers in total, assisting the model in producing the correct answer.
However, it only detects four correctly—missing one and mistakenly counting another one twice.
Finally, \Cref{fig:further_qualitative_examples_15} \textbf{(IV)} illustrates a failure case with LLaVA-OneVision.
Despite access to a high-resolution object relevance map, \texttt{FOCUS} fails to detect the region associated with the umbrella.
As a result, it does not provide the necessary input for the \ac{MLLM} to answer the \ac{VQA} example correctly.

\end{document}